\let\texyear\year
\documentclass{ieeeaccess}
\let\ieeeaccessyear\year
\let\year\texyear
    \usepackage{orcidlink}
    \usepackage{tcolorbox}
    \usepackage{algorithmic}
    \usepackage{graphicx}
    \usepackage{gensymb}
    \usepackage[font={sf,scriptsize}, labelfont={bf,color=accessblue}, caption=false]{subfig}
    \usepackage{colortbl}
    \usepackage{textcomp}
    \usepackage{float}
    \usepackage{hyperref}
\let\year\ieeeaccessyear
\definecolor{accessblue}{RGB}{0,105,154}

\usepackage{cite}
\usepackage{amsmath,amssymb,amsfonts}
\def\BibTeX{{\rm B\kern-.05em{\sc i\kern-.025em b}\kern-.08em
    T\kern-.1667em\lower.7ex\hbox{E}\kern-.125emX}}

\begin{document}
\history{Date of publication xxxx 00, 0000, date of current version xxxx 00, 0000.}
\doi{xxxx}

\title{Gaussian Splatting: 3D Reconstruction and Novel View Synthesis, a Review}
\author{\uppercase{Anurag Dalal}\authorrefmark{\orcidlink{0009-0007-9228-8222}}, \IEEEmembership{Graduate Student, IEEE},
\uppercase{Daniel Hagen}\authorrefmark{\orcidlink{0000-0002-7030-6676}}, \IEEEmembership{Member, IEEE},
\uppercase{Kjell G. Robbersmyr}\authorrefmark{\orcidlink{0000-0001-9578-7325}}, \IEEEmembership{Senior Member, IEEE},
\uppercase{and Kristian Muri Knausgård}\authorrefmark{\orcidlink{0000-0003-4088-1642}}, \IEEEmembership{Senior Member, IEEE}}
\address{Top Research Centre Mechatronics (TRCM), Department of Engineering Sciences, University of Agder, Grimstad, Norway}

\tfootnote{We would like to extend our sincere thanks to Aust Agder utviklings og kompetansefond (AAUKF) for the generous funding of Arven etter Dannevig (The legacy of Dannevig), nr 62/22 which has been instrumental in the completion of this paper. We would also like to thank ChatGPT and Writefull AI platforms for helping us with the writing experience.}

\markboth
{Author \headeretal: Preparation of Papers for IEEE TRANSACTIONS and JOURNALS}
{Author \headeretal: Preparation of Papers for IEEE TRANSACTIONS and JOURNALS}

\corresp{Corresponding author: Anural Dalal (e-mail: anurag.dalal@uia.no).}

\begin{abstract}
Image-based 3D reconstruction is a challenging task that involves inferring the 3D shape of an object or scene from a set of input images. Learning-based methods have gained attention for their ability to directly estimate 3D shapes. This review paper focuses on state-of-the-art techniques for 3D reconstruction, including the generation of novel, unseen views. An overview of recent developments in the Gaussian Splatting method is provided, covering input types, model structures, output representations, and training strategies. Unresolved challenges and future directions are also discussed. Given the rapid progress in this domain and the numerous opportunities for enhancing 3D reconstruction methods, a comprehensive examination of algorithms appears essential. Consequently, this study offers a thorough overview of the latest advancements in Gaussian Splatting.
\end{abstract}


\titlepgskip=-15pt

\maketitle

\section{Introduction}
\label{sec:introduction}
\PARstart{3}{d} reconstruction is a fascinating process that revolves around the creation of three-dimensional models or representations of objects or scenes using 2D images or other data sources~\cite{Samavati2023}. This process aims to transform flat images into immersive and realistic virtual representations that can be utilized in numerous applications. From visualizing architectural designs to animating characters in movies, and from simulating real-world scenarios to analyzing complex structures, 3D reconstruction plays a crucial role in various fields such as computer vision, robotics, and virtual reality. By leveraging advanced algorithms and cutting-edge technologies, researchers and professionals are constantly pushing the boundaries of what is possible in the realm of 3D reconstruction, opening up new possibilities and revolutionizing industries along the way. From the 3D reconstruction of a scene it is possible to render novel viewpoints that are not captured, so this method is called novel view synthesis (NVS). In other words 3D reconstruction enables NVS. In this article, we will dive deeper into the concept of 3D reconstruction and NVS, exploring its methodologies, applications, and the transformative impact it has in our increasingly digital~world.

In recent years, learning-based methods have gained significant prominence, supplanting traditional approaches across various fields of study. These innovative techniques not only offer improved performance, but also introduce novel capabilities. This trend holds true in the realm of 3D computer vision, particularly in 3D reconstruction. For instance, deep learning models have been proposed, enabling end-to-end training and eliminating the need for designing multiple handcrafted stages. Additionally, learning-based methods have the advantage of multitasking, allowing a single model to simultaneously predict both the 3D shape and semantic segmentation of a given scene~\cite{cao2022monoscene}. This integration of advanced learning algorithms has revolutionized the field of 3D reconstruction, offering more efficient and versatile solutions.

Image-based view synthesis techniques play a crucial role in computer graphics and computer vision applications. Addressing the challenge of representing a 3D model or scene based on 2D input images, Gaussian Splatting~\cite{Kerbl2023} emerges as a novel and effective approach. Gaussian Splatting has gained tremendous popularity since it's inception in June 2023 (Figure \ref{fig:gs_trend}). This technique involves iterative refinement of multiple Gaussians to generate 3D objects from 2D images, allowing for the rendering of novel views in complex scenes through interpolation. While Gaussian Splatting doesn't directly recover the entire 3D scene geometry, it stores information in a volumetric point cloud. Each point in this cloud represents a Gaussian with parameters such as color, spread, and location, resulting in a volumetric representation that provides color and density for each point in the relevant 3D space.
\Figure[t!](topskip=0pt, botskip=0pt, midskip=0pt)[width=0.99\columnwidth]{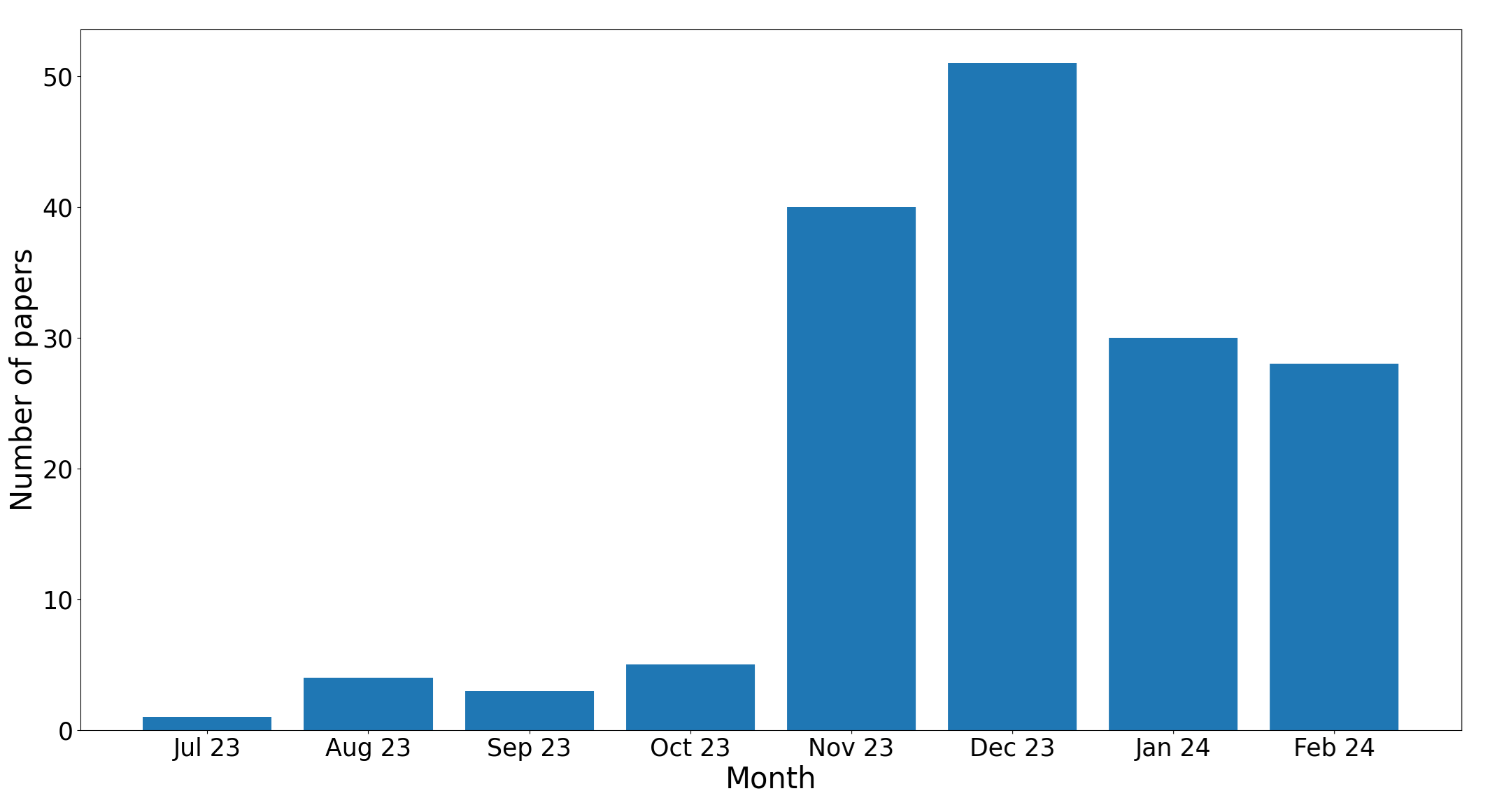}
{ \textbf{Approximate number of publications related to Gaussian Splatting since inception in June 2023 compiled from~\cite{dimensions}.}\label{fig:gs_trend}}

Typically, 3D reconstruction is achieved by traditional algorithms such as photogrammetry and multi view stereo (MVS) algorithms~\cite{Yan2021}. A few modern approaches include Neural Radiance Field (NeRF)~\cite{Mildenhall2020} and Gaussian Splatting~\cite{Kerbl2023}, where Gaussian Splatting is a unique method which is pretty new to the NVS scene. The most common problem faced by any existing solutions is the inability to real time rendering, competitive training time and high quality rendering. Gaussian Splatting offers some significant improvement over NeRFs including fewer artifacts, failure cases, and faster training time. While NeRF already has a few review papers~\cite{Rabby2023, Gao2022}, Gaussian Splatting still don't have a significant review paper that compiles all the recent advancements since its inception.

The objectives of this paper is a thorough review of the various techniques developed in Gaussian Splatting. Section~\ref{sec:sota} presents a comprehensive introduction to the state-of-the-art in 3D reconstruction and NVS and Section~\ref{sec:gaussian_splatting} details the algorithms to achieve 3D reconstruction using Gaussian Splatting.  Section~\ref{sec:fun_adv} deals with the main review part and the latest advancements, and Section~\ref{sec:app_adv} discusses the various applicational areas related to Gaussian Splatting.  Finally, a discussion of the methods, research directions, and conclusion is presented.
\section{A primer on 3D Reconstruction and Novel View Synthesis}
\label{sec:sota}
3D reconstruction and NVS are two closely related fields in computer graphics that aim to capture and render realistic 3D representations of physical scenes. 3D reconstruction involves extracting the geometric and appearance information from a series of 2D images, typically captured from different viewpoints. Although there are numerous techniques for 3D scanning, this capturing of different 2D images is very straightforward and computationally cheap way to gather information about the 3D environment. This information can then be used to create a 3D model of the scene, which can be used for various purposes, such as virtual reality (VR) applications, augmented reality (AR) overlays, or computer-aided design (CAD) modeling.

On the other hand, NVS focuses on generating new 2D views of a scene from a previously acquired 3D model. This allows the creation of realistic images of a scene from any desired viewpoint, even if the original images were not taken from that angle. Recent advances in deep learning have led to significant improvements in both 3D reconstruction and NVS. Deep learning models can be used to efficiently extract 3D geometry and appearance from images, and such models can also be used to generate realistic novel views from 3D models. As a result, these techniques are becoming increasingly popular in a variety of applications, and they are expected to play an even more important role in the future.

This section will introduce how 3D data is stored or represented, followed by the most commonly used publicly available dataset for this task, and then will expand on various algorithms, primarily focusing on Gaussian Splatting.

\subsection{3D Data Representation}
The intricate spatial nature of 3D data, which includes volumetric dimensions, provides a detailed representation of objects and environments. This is crucial for creating immersive simulations and accurate models in various fields of study. The multidimensional structure of 3D data allows the incorporation of depth, width, and height, leading to significant advancements in disciplines such as architectural design and medical imaging techniques.

The selection of the data representation plays a crucial role in the design of numerous 3D deep learning systems. Point clouds, which lack grid-like structures, typically cannot be directly subjected to convolutions. On the other hand, voxel representations, characterized by grid-like structures, often incur high computational memory demands.

The evolution of 3D representation is accompanied by the way in which 3D data or models are stored. The most frequently used representation of 3D data can be classified as traditional and novel approaches:

\subsubsection{Traditional Approaches}

\textbf{Point cloud}: A 3D point cloud~\cite{ma20223d} provides a direct and uncomplicated representation of 3D objects. In this representation, each point cloud consists of a set of 3D points, with each individual point represented by a three-dimensional tuple (x, y, z). Typically, the raw data captured by numerous depth cameras is presented in the form of 3D point clouds.\\
\textbf{Mesh}: Meshes~\cite{ma20223d} serve as another commonly utilized 3D data representation. Similar to the points in point clouds, each mesh comprises a collection of 3D points known as vertices. Additionally, meshes include a set of polygons, referred to as faces, which are defined based on these vertices. In many data-driven applications, meshes result from post-processing raw measurements obtained from depth cameras~\cite{liao2009modeling}. Frequently, they are manually crafted during the creation of 3D assets. In contrast to point clouds, meshes offer extra geometric details, encode topology, and incorporate surface-normal information. This supplementary information proves particularly valuable in the training of learning models. For instance, graph convolutional neural networks often treat meshes as graphs and establish convolutional operations using information about vertex neighbors. \\
\textbf{Voxel}: Yet another crucial representation for 3D data is the voxel representation~\cite{ma20223d}. In 3D computer vision, a voxel is analogous to a pixel. Just as a pixel is delineated by subdividing a 2D rectangle into smaller rectangles, each of which is a pixel, a voxel is defined by partitioning a 3D volume into smaller cubes, with each individual cuboid referred to as a voxel. Typically, voxel representations employ truncated signed distance functions (TSDFs) for the portrayal of 3D surfaces. At each voxel, a signed distance function (SDF) can be established as the (signed) distance from the center of the voxel to the nearest point on the surface. A positive sign in the SDF signifies that the center of the voxel lies outside an object. The key distinction between a TSDF and an SDF lies in the truncation of values; TSDF values are truncated, ensuring that they fall consistently within the range of -1~to~+1.
Figure~\ref{fig:traditional_data} shows how 3D data are represented in traditional techniques.
\Figure[hbt!](topskip=0pt, botskip=0pt, midskip=0pt)[width=.4\columnwidth]{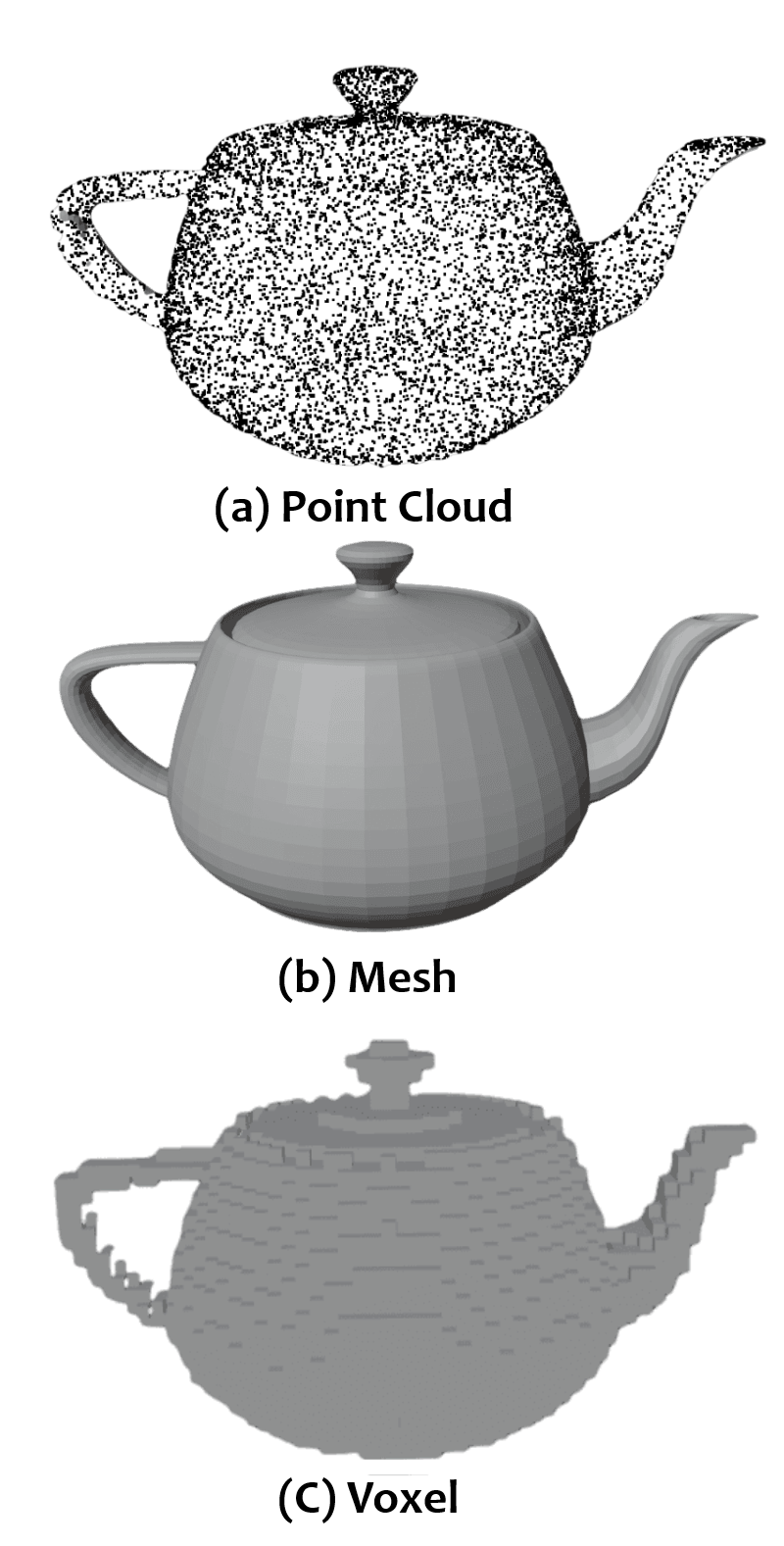}
{ \textbf{Traditional 3D data representations. \textit{3D model source: }\cite{teapot}.}\label{fig:traditional_data}}
\subsubsection{Novel Approaches}
\textbf{Neural Network/Multi layer perceptron (MLP)}: NeRFs~\cite{Mildenhall2020} are a type of 3D deep learning model that can represent and render 3D scenes with high fidelity. They encode 3D information in a unique way that combines traditional 3D geometry with a neural network representation. To capture more intricate details and lighting effects, NeRFs augment the density field with a neural network. This network, known as the radiance field, takes a 3D position as input and outputs a color value along with a normal vector. The color represents the color of the surface at the given position, while the normal vector indicates the surface's orientation.
The radiance field is trained on a dataset of images, allowing it to learn how to map 3D positions to their corresponding color and normal information. This information is crucial for generating photorealistic images from arbitrary viewpoints.\\
\textbf{Gaussian Splats}: Gaussian splats~\cite{Kerbl2023} are a type of 3D representation that is used to render complex scenes with high fidelity. They are a more efficient and flexible alternative to traditional methods such as point clouds or voxel grids. 
Gaussian splats are stored in a compact format that represents each splat as a collection of parameters. These parameters typically include: 
\begin{itemize}
\item Position: The 3D location of the center of the splat. 
\item Scale: The size of the splat. 
\item Opacity: The amount of influence the splat has on the rendered image. 
\item Color: The color of the splat. 
\item Material properties: Additional properties such as shininess, reflection, and refraction.
\end{itemize}
By storing splats in this compact format, it is possible to represent a large number of splats with relatively less data. This makes Gaussian splats well-suited for rendering complex scenes at high resolutions.
Figure~\ref{fig:novel_data} shows how 3D data are represented in these two novel techniques.
\Figure[t!](topskip=0pt, botskip=0pt, midskip=0pt)[width=0.99\columnwidth]{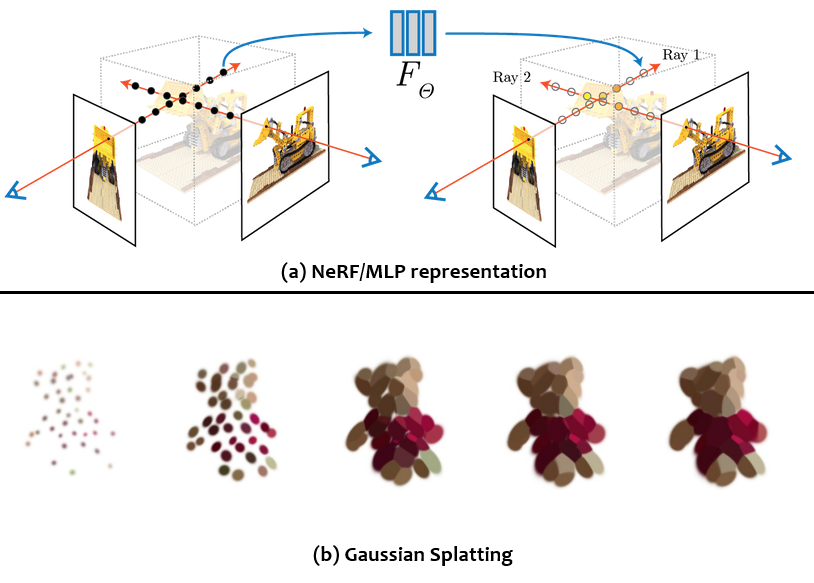}
{\textbf{Novel 3D data representations~\cite{Mildenhall2020, Kerbl2023}.}\label{fig:novel_data}}
\subsection{Datasets}
Gaussian Splatting models are generally modeled per scene and demand dense images of a scene with a varied number of poses. In most cases, the camera poses are unknown and are calculated using the structure from motion (SFM)~\cite{schoenberger2016sfm} using the COLMAP library~\cite{COLMAP}.
The original Gaussian Splatting paper uses three datasets: the Mip-NeRF360~\cite{barron2022mipnerf360}, Tanks\&Temples~\cite{Knapitsch2017}, and Deep Blending~\cite{DeepBlending2018}.

As research progresses in 3D reconstruction, MVS, and NVS, an increasing number of datasets, are becoming available for testing the algorithms developed in these areas. Some of the notable once are listed in Table~\ref{table:dataset_table}.
\begin{table*}[hbt!]
\centering
\caption{Most commonly used 3D reconstruction and NVS datasets, sorted based on ranking from Papers with code. The last eight entries are taken based on the algorithms that are dicussed but not present in Papers with code ranking.}
\label{table:dataset_table}
\begin{tabular}{|>{\centering\arraybackslash}m{57pt}|m{280pt}|m{127pt}|}
\hline
\textbf{Dataset Name} & \textbf{Description} & \textbf{Source} \\
\hline
LLFF & Local Light Field Fusion (LLFF) dataset includes synthetic and real images of natural scenes. Synthetic images are generated from SUNCG and UnrealCV, while real images consist of 24 scenes captured using a handheld cellphone. & \url{https://bmild.github.io/llff/}~\cite{mildenhall2019llff} \\
\hline
NeRF & Neural Radiance Fields (NeRF)  dataset comprises synthetic renderings and real images of complex scenes. It includes Diffuse Synthetic 360 \textdegree, Realistic Synthetic 360\textdegree, and real images of complex scenes. & \url{https://www.matthewtancik.com/nerf}~\cite{mildenhall2020nerf} \\
\hline
DONeRF & DONeRF dataset incorporates synthetic data generated using Blender and Cycles path tracer, with 300 images rendered for each scene. & \url{https://depthoraclenerf.github.io/}~\cite{Neff_2021} \\
\hline
X3D & X3D dataset comprises 15 scenes dedicated to X-ray 3D reconstruction, spanning medicine, biology, security, and industry applications. & \url{https://github.com/caiyuanhao1998/SAX-NeRF}~\cite{cai2023structureaware} \\
\hline
RTMV & RTMV is a synthetic dataset for novel view synthesis, consisting of 300,000 images generated through ray tracing across 2,000 scenes. & \url{https://www.cs.umd.edu/~mmeshry/projects/rtmv/}~\cite{tremblay2022rtmv} \\
\hline
Tanks\&Temples & Tanks\&Temples dataset is comprehensive, featuring both intermediate and advanced testing datasets for image-based 3D reconstruction pipelines. & \url{https://www.tanksandtemples.org/}~\cite{Knapitsch2017} \\
\hline
RealEstate10K & RealEstate10K is a large dataset of camera poses derived from 10,000 YouTube videos, providing trajectories obtained through SLAM and bundle adjustment algorithms. & \url{https://google.github.io/realestate10k/}~\cite{RealEstate10K} \\
\hline
ACID & Aerial Coastline Imagery Dataset (ACID) dataset focuses on generating novel views over an extended camera trajectory based on a single image, using a hybrid approach of geometry and image synthesis. & \url{https://infinite-nature.github.io/}~\cite{infinite_nature_2020} \\
\hline
SWORD & 'Scenes with occluded regions' dataset (SWORD) dataset comprises 1,500 training videos and 290 test videos, emphasizing nearby objects and occlusions for robust model training. & \url{https://samsunglabs.github.io/StereoLayers/}~\cite{Khakhulin2021SLayers} \\
\hline
Mip-NeRF 360 & Mip-NeRF 360 dataset extends Mip-NeRF with non-linear parameterization, online distillation, and a distortion-based regularizer for unbounded scenes. & \url{https://jonbarron.info/mipnerf360/}~\cite{barron2022mipnerf360} \\
\hline
Deep Blending & Deep Blending dataset for Free-Viewpoint Image-Based Rendering includes 9 scenes captured with a stereo camera rig and reconstructed using COLMAP and RealityCapture. & \url{http://visual.cs.ucl.ac.uk/pubs/deepblending/}~\cite{DeepBlending2018} \\
\hline
DTU & DTU dataset is multi-view stereo data with precise camera positioning, structured light scanner, and diverse scenes with varying illumination. & \url{https://roboimagedata.compute.dtu.dk/?page_id=36}~\cite{jensen2014large} \\
\hline
ScanNet & ScanNet is an indoor RGB-D dataset with 1513 annotated scans, providing 90\% surface coverage and diverse 3D scene understanding tasks. & \url{http://www.scan-net.org/}~\cite{dai2017scannet} \\
\hline
ShapeNet & ShapeNet is a large-scale repository for 3D CAD models, valuable for NeRF models emphasizing object-based semantic labels. & \url{https://shapenet.org/}~\cite{chang2015shapenet} \\
\hline
Matterport 3D & Matterport-3D dataset includes 10,800 panoramic views from 90 building-scale scenes with depth, semantics, and instance annotations. & \url{https://niessner.github.io/Matterport/}~\cite{Matterport3D} \\
\hline
Replica & Replica dataset is a genuine indoor dataset with 18 scenes and 35 rooms, featuring manual adjustments, semantic annotations, and both class-based and instance-based labels. & \url{https://github.com/facebookresearch/Replica-Dataset}~\cite{replica19arxiv} \\
\hline
Plenoptic Video & Plenoptic Video dataset comprises 3D videos captured using a plenoptic camera for realistic and immersive 3D experiences. & \url{https://neural-3d-video.github.io/}~\cite{labussiere2020blur_plenoptic} \\
\hline
Panoptic & CMU Panoptic dataset features 3D pose annotations for over 1.5 million instances in social activities, captured with synchronized cameras and diverse scenes. & \url{http://domedb.perception.cs.cmu.edu/}~\cite{Joo_2017_TPAMI_panoptic} \\
\hline
\end{tabular}
\end{table*}
\subsection{3D reconstruction and NVS techniques}

To evaluate the current advances in the field, a literature study is conducted, identifying and scrutinizing relevant academic works. The analysis specifically concentrates on two key areas: 3D reconstruction and NVS.
The evolution of 3D volume reconstruction from multiple camera images spans several decades and has diverse applications in computer graphics, robotics, and medical imaging. The state of the art is explored in the next part.

\textbf{Photogrammetry}: Since the 1980s, advanced photogrammetry and stereo vision techniques emerged, automating the identification of corresponding points in stereo image pairs~\cite{Deliry2021}. Photogrammetry is a method merging photography and computer vision to generate 3D models of objects or scenes. It entails capturing images from various perspectives, utilizing software like Agisoft Metashape~\cite{agisoft} to estimate camera positions and generate a point cloud. This point cloud is subsequently transformed into a textured 3D mesh, enabling the creation of detailed and photorealistic visualizations of the reconstructed object or scene.

\textbf{Structure from motion}: In the 1990s, SFM techniques gained prominence, enabling the reconstruction of 3D structure and camera motion from sequences of 2D images~\cite{Deliry2021}. SFM is the process of estimating the 3D structure of a scene from a set of 2D images. SFM requires point correspondences between images. Finding corresponding points either by matching features or tracking points from multiple images, and triangulating to find 3D positions.

\textbf{Deep learning}: Recent years have seen the integration of deep learning techniques, particularly Convolutional Neural Networks (CNNs)~\cite{Yan2021}. Deep learning based methods have picked up pace in 3D reconstruction. The most notable is 3D Occupancy Network, a type of neural network architecture designed for 3D scene understanding and reconstruction~\cite{Huang2023, Peng2020}. It operates by dividing a 3D space into small volumetric cells or voxels, with each voxel representing whether it contains objects or is empty space. These networks use deep learning techniques, like 3D convolutional neural networks, to predict voxel-wise occupancy, making them valuable for applications such as robotics, autonomous vehicles, augmented reality, and 3D scene reconstruction. These networks heavily rely on convolution and transformers~\cite{Zhang2023, Mao2023, Huang2023}. They are essential for tasks like collision avoidance, path planning, and real-time interaction with the physical world. Additionally, 3D Occupancy Networks can estimate uncertainty, but they may have computational limitations in handling dynamic or complex scenes. Advances in neural network architectures continue to improve their accuracy and efficiency.

\textbf{Neural Radiance Field}: NeRF was introduced in 2020~\cite{Mildenhall2020}, and it integrates neural networks with classical 3D reconstruction principles, gaining notable attention in computer vision and graphics~\cite{ Zhang2022}. It reconstructs detailed 3D scenes by modeling volumetric functions, predicting color and density through a neural network. NeRFs are widely applied in computer graphics, and virtual reality. Recently, NeRF has seen enhancements in accuracy and efficiency through extensive research~\cite{Rabby2023, Gao2022}. Recent studies also explore NeRF's applicability in underwater scenes~\cite{Sethuraman2022}. While offering a robust representation of 3D scene geometry still challenges like computational demands persist. Future NeRF research needs to focus on interpretability, real-time rendering, novel applications, and scalability, opening avenues in virtual reality, gaming, and robotics~\cite{Mildenhall2020, Rabby2023}.

\textbf{Gaussian Splatting}: Finally, in 2023  3D Gaussian Splatting~\cite{Kerbl2023} emerged as a novel technique for real-time 3D rendering. In the next section this method is discussed in details.

\section{Fundamentals of Gaussian Splatting}
\label{sec:gaussian_splatting}
Gaussian Splatting portrays a 3D scene using numerous 3D Gaussians or particles, each equipped with position, orientation, scale, opacity, and color information. To render these particles, they undergo a conversion to 2D space and are strategically organized for optimal rendering. 

Figure~\ref{fig:gaussian_splatting_algo} shows the architecture of the Gaussian Splatting algorithm. In the original algorithm, the following steps are taken:

\begin{enumerate}
\item\textbf{Structure from motion:} The process starts using the SFM method~\cite{schoenberger2016sfm} to create a point cloud from images using the COLMAP library~\cite{COLMAP}.
\item\textbf{Convert to gaussian splats:} The conversion of each point to Gaussian Splats enables rasterization. The SFM data only allows for the initialization of position, size, and color of each splats.
\item\textbf{Training:} To ensure a representation yields high-quality outcomes, training is imperative. Stochastic Gradient Descent, akin to a neural network, is employed for this purpose.
\begin{itemize}
\item Employ differentiable Gaussian rasterization to rasterize Gaussians in an image.
\item Compute the loss based on the disparity between the raster and actual terrain images.
\item Modify the Gaussian parameters in accordance with the incurred loss.
\end{itemize}
\item\textbf{Differentiable Gaussian rasterization:} Every 2D Gaussian requires differentiable Gaussian rasterization to be projected from the viewpoint of the camera, sorted according to depth, then repeated both backwards and forwards combined for every pixel. A detailed mathematical insight to the rasterization and training process can be found in~\cite{Ye2023supplement}.
\end{enumerate}
\Figure[hbt!](topskip=0pt, botskip=0pt, midskip=0pt)[scale=0.8] {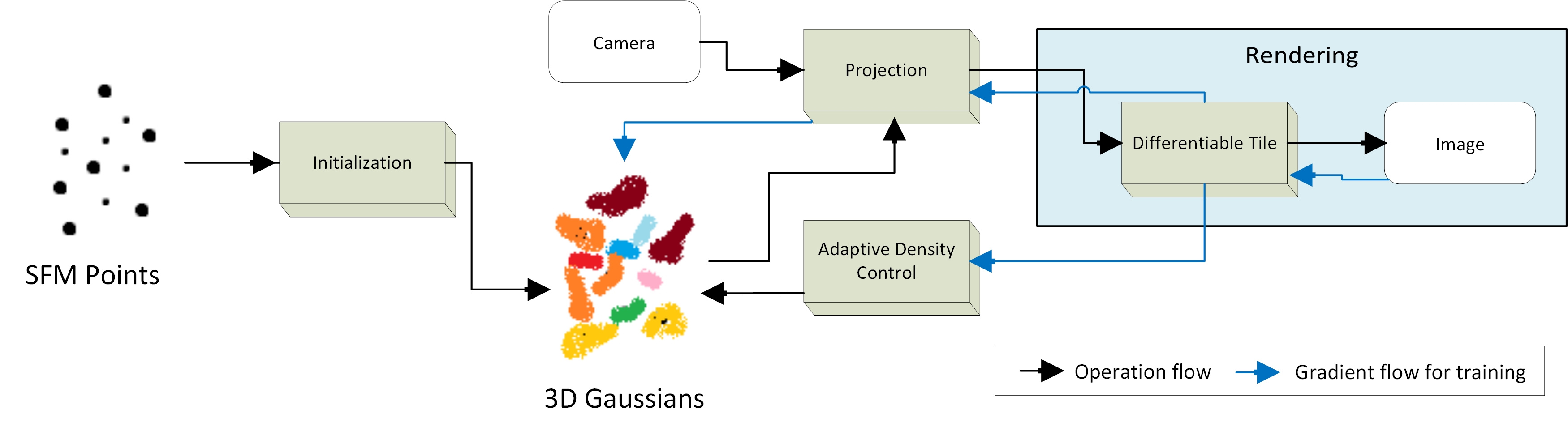}
{ \textbf{3D Gaussian Splatting architecture~\cite{Kerbl2023}.}
\label{fig:gaussian_splatting_algo}}
\subsection{Mathematical Representation and Rendering process}
The mathematical insights discussed in this section is formalized by Ye et al. in~\cite{Ye2023}.
A 3D Gaussian is parameterized by its mean $\mu \in \mathbb{R}^3$,
covariance $\Sigma \in \mathbb{R}^{3\times 3}$,
color $c \in \mathbb{R}^3$, and opacity $o \in \mathbb{R}$.
To render a view of the Gaussians, first compute their projected 2D locations and extents in the camera plane. The visible 2D Gaussians are then sorted by depth and composited from front to back to construct the output image.
\subsubsection{Projection of Gaussians}
The render camera is described by its extrinsics $T_{\textrm{cw}}$,
which transforms points from the world coordinate space to the camera coordinate space,
and its intrinsics, which are the focal length $(f_x, f_y)$ and the principal point $(c_x, c_y)$ of the camera plane.
The transformation from camera space to normalized clip space is denoted with the projection matrix $P$.
\begin{equation}
T_{\textrm{cw}}  = \begin{bmatrix}
    R_{\textrm{cw}} & t_{\textrm{cw}}\\
    0 & 1
\end{bmatrix} \in SE(3),
\end{equation}
\begin{equation}
{
\quad
P = \begin{bmatrix}
\frac{2f_x}{w} & 0 & 0 & 0\\
0 & \frac{2f_y}{h} & 0 & 0 \\
0 & 0 & \frac{f+n}{f-n} & \frac{-2fn}{f-n}\\
0 & 0 & 1 & 0
\end{bmatrix},}
\end{equation}
where $(w, h)$ are the output image width and height, and $(n, f)$ are the near and far clipping planes.
The 3D mean $\mu$ is projected into pixel space via standard perspective projection.
This is achieved by transforming the mean $\mu$ into $t \in \mathbb{R}^4$ in camera coordinates,
$t' \in \mathbb{R}^4$ in ND coordinates, and 
$\mu' \in \mathbb{R}^2$ in pixel coordinates
\begin{equation}
{
    t = T_{\textrm{cw}} \begin{bmatrix} \mu \\ 1 \end{bmatrix},
    \quad t' = P t,
    \quad \mu' = \begin{bmatrix}
        \frac{1}{2}(\frac{w t'_x}{t'_w} + 1) + c_x \\
        \frac{1}{2}(\frac{h t'_y}{t'_w} + 1) + c_y \\
\end{bmatrix}}
\end{equation}

Perspective projection of a 3D Gaussian does not result in a 2D Gaussian.
The projection of $\Sigma$ to pixel space is approximated with a first-order Taylor expansion at $t$ in the camera frame.
Specifically, the affine transform is computed as
$J \in \mathbb{R}^{2\times 3}$ as shown in~\cite{ewasplatting} as
\begin{equation}
    J = \begin{bmatrix}
    \frac{f_x}{t_z} & 0 & -\frac{f_x t_x}{t_z^2} \\
    0 & \frac{f_y}{t_z} & -\frac{f_y t_y}{t_z^2} \\
    \end{bmatrix}
\end{equation}

The transformed 2D covariance matrix $\Sigma' \in \mathbb{R}^{2\times 2}$ is then given by
\begin{equation}
    \Sigma' = J R_{\textrm{cw}} \Sigma R_{\textrm{cw}}^\top J^\top
\end{equation}

Finally, the 3D covariance $\Sigma$ with scale $s \in \mathbb{R}^3$
and rotation quaternion $ q \in \mathbb{R}^4$ is parameterized and converted to $\Sigma$.
Then the quaternion $q = (x, y, z, w)$ is converted into a rotation matrix:
{\small
\begin{equation}
R =  \begin{bmatrix}
        1 - 2  (y^2 + z^2) & 2  (x y - w z) & 2  (x z + w y)\\
        2  (x y + w z) & 1 - 2  (x^2 - z^2) & 2  (y z - w x)\\
        2  (x z - w y) & 2  (y z + w x) & 1 - 2  (x^2 + y^2)
    \end{bmatrix}
\end{equation}}
The 3D covariance $\Sigma$ is then given by 
\begin{equation}
    \Sigma = RS S^\top R^\top,
\end{equation}
where $S = \mathrm{diag}(s) \in \mathbb{R}^{3\times 3}$

\subsubsection{Depth Compositing of Gaussians}
For every Gaussian, the bounding box aligned with the axis encompassing the 99\% confidence ellipse of each 2D projected covariance ($3 \sigma$) is calculated. If the bounding box intersects with the tile, the Gaussian is added to the respective tile bin. Subsequently, the authors implement the tile sorting algorithm outlined in~\cite{Kerbl2023}, Appendix C to generate a sorted list of Gaussians based on depth for each tile.
Then the sorted Gaussians within each tile is rasterized.
For a color at a pixel $i$, let $n$ index the $N$ Gaussians involved in that pixel, it is calculated as:
\begin{equation}
    C_i = \sum_{n \le N} c_n \alpha_n T_n, ~~\textrm{where}~~ T_n = \prod_{m < n} (1 - \alpha_m)
\end{equation}
And, the opacity $\alpha$ with the 2D covariance $\Sigma' \in \mathbb{R}^{2\times 2}$ calculated as:
\begin{align*}
    \alpha_n = o_n \exp(-\sigma_n), \quad
    \sigma_n = \frac{1}{2} \Delta_n^\top \Sigma'^{-1} \Delta_n
\end{align*}
where $\Delta \in \mathbb{R}^2$ and is the offset between the pixel center and the 2D Gaussian center $\mu' \in \mathbb{R}^2$.
$T_n$ is computed online as the interaction is made through the Gaussians from front to back.

\subsection{Quality Assessment Matrices}
In the conventional scenario of Gaussian Splatting, benchmarking for NVS often involves the use of visual quality assessment metrics. These metrics aim to evaluate the quality of individual images, either with (full-reference) or without (no-reference) ground truth images. The peak signal-to-noise ratio (PSNR), structural similarity index measure (SSIM)~\cite{SSIM1284395}, and learned perceptual image patch similarity (LPIPS)~\cite{LPIPSzhang2018unreasonable} are widely utilized in the Gaussian Splatting literature as the primary metrics for this purpose. The mathematical formulation of theses metrices are defined below.
\subsubsection{Peak Signal to Noise Ratio}
PSNR↑ is a no-reference quality assessment metric. PSNR is defined by the following formulae:\\
\begin{equation}
PSNR(I) = 10 \cdot \log_{10} \left(\frac{MAX(I)^2}{MSE(I)}\right),
\end{equation}
where \begin{math}MAX(I)\end{math} is the maximum possible pixel value in
the image (255 for 8-bit integer), and \begin{math}MSE(I)\end{math} is the pixel-wise mean squared error calculated on all color channels. PSNR is also commonly used in signal processing and is well understood.
\subsubsection{Structural Similarity Index Measure}
SSIM↑~\cite{SSIM1284395} is a full-reference quality assessment metric.The SSIM for a single patch is given by the following formula:
\begin{equation}S S I M(x,y)=\frac{(2\mu_{x}\mu_{y}+C_{1})(2\sigma_{x y}+C_{2})}{(\mu_{x}^{2}+\mu_{y}^{2}+C_{1})(\sigma_{x}^{2}+\sigma_{y}^{2}+C_{2})},\end{equation}
where \begin{math}C_{i} = (K_{i}L)^2\end{math}, L is the dynamic range of the pixels (255 for 8bit integer), and \begin{math} K_{1} = 0.01, K_{2} = 0.03\end{math} are constants chosen in~\cite{SSIM1284395}. The local statistics \begin{math}\mu^{\prime}s,\sigma^{\prime}s\end{math} are calculated within a \begin{math}11 \times 11\end{math} circular symmetric Gaussian weighted window, with weights \begin{math}w_{i}\end{math} having a standard deviation of 1.5 and normalized to 1.
\subsubsection{Learned Perceptual Image Patch Similarity}
LPIPS↓~\cite{LPIPSzhang2018unreasonable} is a complete reference quality assessment metric that uses learned convolutional characteristics. The score is given by a weighted pixel-wise MSE of feature maps over multiple layers.
\begin{equation}L P I P S(x,y)=\sum_{l}^{L}\frac{1}{H_{l}W_{l}}\sum_{h.w}^{H_{l},W_{l}}||w_{l}(\varsigma)(x_{h w}^{l}-y_{h w}^{l})||_{2}^{2},\end{equation}
where \begin{math}x_{hw}^l, y_{hw}^l\end{math} are the original and generated/accessed images’ feature at pixel width \begin{math}w\end{math}, pixel height \begin{math}h\end{math}, and layer \begin{math}l\end{math}.  \begin{math}W_{l}\end{math} and \begin{math}H_{l}\end{math} are the width and height of feature map at the corresponding layer. The original LPIPS paper used SqueezeNet~\cite{iandola2016squeezenet}, VGG~\cite{simonyan2015deepvgg} and AlexNet~\cite{NIPS2012_c399862d_alex} as feature extraction backbone.
\subsection{State of Art}
\label{sec:adv}
In the next two sections, an exploration of various applications and advancements in Gaussian Splatting will be undertaken, delving into its diverse implementations across domains such as autonomous driving, avatars, compression, diffusion, dynamics and deformation, editing, text-based generation, mesh extraction and physics, regularization and optimization, rendering, sparse representations, and simultaneous localization and mapping (SLAM). Each subcategory will be examined to provide insights into the versatile use of Gaussian Splatting methodologies in addressing specific challenges and achieving notable advancements within these distinct domains. Figure~\ref{fig:taxonomy_innovations} shows complete list of all the methods that are discussed in Section~\ref{sec:fun_adv} and~\ref{sec:app_adv}. Broadly, the division of methods can be classified into functional areas based on functional improvements such as compression, rendering in Section~\ref{sec:fun_adv} and Specific use cases are applicational areas such as avatars, SLAM in Section~\ref{sec:app_adv}.
\definecolor{color1}{HTML}{00263d}
\definecolor{color2}{HTML}{004873}
\definecolor{color3}{HTML}{005f98}
\definecolor{color4}{HTML}{1883c4}
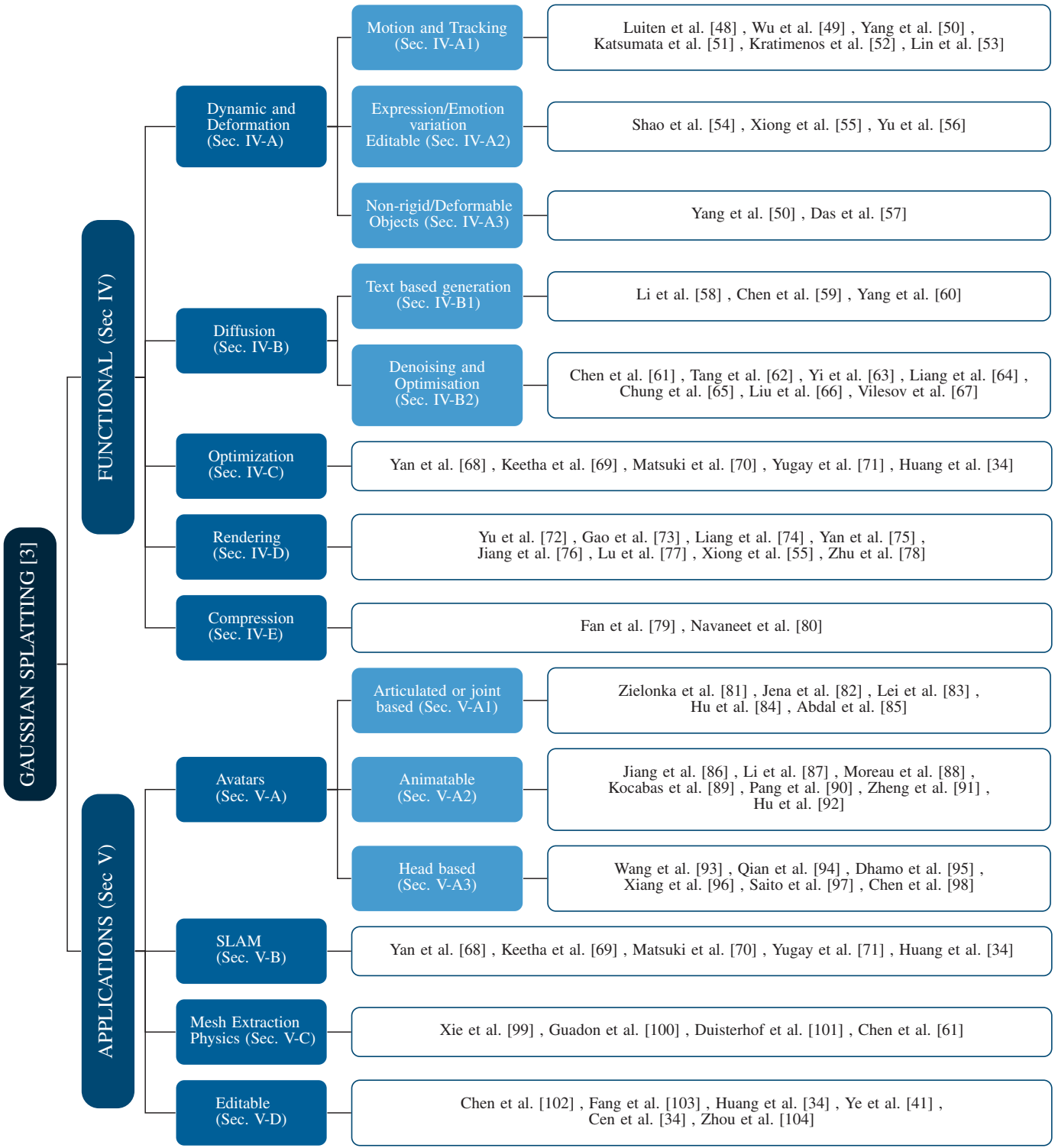
\begin{figure*}[htbp!]
 \definecolor{hiddendraw}{rgb}{0.5, 0.5, 0.5}  
 \centering
\begin{forest}
  forked edges,
  for tree={
    grow=east,
    reversed=true,
    anchor=base west,
    parent anchor=east,
    child anchor=west,
    base=left,
    font=\small,
    rectangle,
    draw={hiddendraw, line width=0.6pt},
    rounded corners,align=left,
    minimum width=2.5em,
    edge={black, line width=0.55pt},
    l+=1.9mm,
    s sep=7pt,
    inner xsep=7pt,
    inner ysep=8pt,
    ver/.style={rotate=90, rectangle, draw=none, rounded corners=3mm, fill=color1, text centered,  text=white, child anchor=north, parent anchor=south, anchor=center, font=\fontsize{10}{10}\selectfont,},
    ver2/.style={rotate=90, rectangle, draw=none, rounded corners=3mm, fill=color2, text centered,  text=white, child anchor=north, parent anchor=south, anchor=center, text width = 14em, font=\fontsize{10}{10}\selectfont,},
    level2/.style={rectangle, draw=none, fill=color3,  
    text centered, anchor=west, text=white, font=\fontsize{8}{8}\selectfont, text width = 6em},
    level3/.style={rectangle, draw=none, fill=color4,   fill opacity=0.8,text centered, anchor=west, text=white, font=\fontsize{8}{8}\selectfont, text width = 7em, align=center},
    level3_2/.style={rectangle, draw=none, fill=gray,   fill opacity=0.8,text centered, anchor=west, text=white, font=\fontsize{8}{8}\selectfont, text width = 3.0cm, align=center},
    level3_1/.style={rectangle, draw=none, fill=brown,   fill opacity=0.8, text centered, anchor=west, text=white, font=\fontsize{8}{8}\selectfont, text width = 3.2cm, align=center},
    level4/.style={rectangle, draw=color2, text centered, anchor=west, text=black, font=\fontsize{8}{8}\selectfont, align=center, text width = 2.9cm},
    level5/.style={rectangle, draw=color2, text centered, anchor=west, text=black, font=\fontsize{8}{8}\selectfont, align=center, text width = 14.7em},
    level5_1/.style={rectangle, draw=color2, text centered, anchor=west, text=black, font=\fontsize{8}{8}\selectfont, align=center, text width = 23.3em},
    level5_2/.style={rectangle, draw=color2, text centered, anchor=west, text=black, font=\fontsize{8}{8}\selectfont, align=center, text width = 33.0em},
  },
  where level=1{text width=5em,font=\scriptsize,align=center}{},
  where level=2{text width=6em,font=\tiny,}{},
  where level=3{text width=6em,font=\tiny}{},
  where level=4{text width=5em,font=\tiny}{},
  where level=5{font=\tiny}{},
  [GAUSSIAN SPLATTING \cite{Kerbl2023}, ver
    [FUNCTIONAL (Sec \ref{sec:fun_adv}), ver2
        [Dynamic and \\Deformation \\(Sec. \ref{tree:1}), level2
            [Motion and Tracking \\(Sec. \ref{tree:1_1}), level3
                    [Luiten et al.\cite{Luiten2023} {,} 
                     Wu et al.\cite{Wu2023} {,} 
                     Yang et al.\cite{yang2023deformable3dgs} {,}\\ 
                     Katsumata et al.\cite{Katsumata2023} {,} 
                     Kratimenos et al.\cite{Kratimenos2023} {,} 
                     Lin et al.\cite{Lin2023} 
                     , level5_1]]
            [Expression/Emotion \\ variation\\Editable (Sec. \ref{tree:1_2}), level3
                    [Shao et al.\cite{Shao2023} {,} 
                     Xiong et al.\cite{Xiong2023_SparseGS} {,} 
                     Yu et al.\cite{Yu2023} 
                    , level5_1]]
            [Non-rigid/Deformable \\ Objects (Sec. \ref{tree:1_3}), level3
                    [Yang et al.\cite{yang2023deformable3dgs} {,} 
                     Das et al.\cite{Das2023} 
                    , level5_1]]
        ]
        [Diffusion\\ (Sec. \ref{tree:2}), level2
            [Text based generation\\(Sec. \ref{tree:2_1}), level3
                    [Li et al.\cite{Li2023} {,} 
                     Chen et al.\cite{Chen2023_text} {,}
                     Yang et al.\cite{Yang2023prior} 
                    , level5_1]
            ]
            [Denoising and \\Optimisation \\(Sec. \ref{tree:2_2}), level3
                    [Chen et al.\cite{Chen2023} {,} 
                     Tang et al.\cite{Tang2023} {,} 
                     Yi et al.\cite{Yi2023} {,} 
                     Liang et al.\cite{Liang2023} {,}\\ 
                     Chung et al.\cite{Chung2023} {,} 
                     Liu et al.\cite{liu2023HumanGaussian} {,} 
                     Vilesov et al.\cite{Vilesov2023} 
                    , level5_1]
            ]
        ]
        [Optimization\\(Sec. \ref{tree:6}), level2
            [Yan et al. \cite{yan2023nerfds} {,}
             Keetha et al. \cite{keetha2023splatam} {,}
             Matsuki et al. \cite{matsuki2023gaussian} {,}
             Yugay et al. \cite{Yugay2023Gaussian-SLAM} {,}
             Huang et al. \cite{Huang2023}
             ,level5_2]
        ]
        [Rendering\\(Sec. \ref{tree:8}), level2
            [Yu et al. \cite{Yu2023Mip_Splatting} {,}
             Gao et al. \cite{Gao2023} {,}
             Liang et al. \cite{Liang2023GS_IR} {,}
             Yan et al. \cite{Yan2023Multi_Scale} {,}\\
             Jiang et al. \cite{Jiang2023_3d_gaussian} {,}
             Lu et al. \cite{Lu2023} {,}
             Xiong et al. \cite{Xiong2023_SparseGS} {,}
             Zhu et al. \cite{Zhu2023}
             ,level5_2]
        ]
        [Compression\\(Sec. \ref{tree:9}), level2
            [Fan et al. \cite{Fan2023} {,}
             Navaneet et al. \cite{Navaneet2023}
             ,level5_2]
        ]]
    [APPLICATIONS  (Sec \ref{sec:app_adv}), ver2
        [Avatars \\ (Sec. \ref{tree:3}), level2
            [Articulated or joint\\ based (Sec. \ref{tree:3_1}), level3
                [Zielonka et al. \cite{Zielonka2023} {,}
                 Jena et al. \cite{Jena2023} {,}
    			 Lei et al. \cite{Lei2023} {,}\\
    			 Hu et al. \cite{Hu2023gauhuman} {,}
                 Abdal et al. \cite{Abdal2023shell}
                , level5_1]
            ]
            [Animatable\\(Sec. \ref{tree:3_2}), level3
                [Jiang et al. \cite{Jiang2023} {,}
                 Li et al. \cite{Li2023animatable} {,}
    			 Moreau et al. \cite{Moreau2023} {,}\\
    			 Kocabas et al. \cite{Kocabas2023} {,}
    			 Pang et al. \cite{Pang2023} {,}
    			 Zheng et al. \cite{Zheng2023} {,}\\
                 Hu et al. \cite{Hu2023GaussianAvatar}
                , level5_1]
            ]
            [Head based\\(Sec. \ref{tree:3_3}), level3
                [Wang et al. \cite{Wang2023} {,}
    			 Qian et al. \cite{Qian2023} {,}
    			 Dhamo et al. \cite{Dhamo2023} {,}\\
    			 Xiang et al. \cite{Xiang2023} {,}
    			 Saito et al. \cite{Saito2023} {,}
    			 Chen et al. \cite{Chen2023mono}
                 ,level5_1]
            ]
        ]
        [SLAM\\(Sec. \ref{tree:4}), level2
            [Yan et al. \cite{yan2023nerfds} {,}
             Keetha et al. \cite{keetha2023splatam} {,}
             Matsuki et al. \cite{matsuki2023gaussian} {,}
             Yugay et al. \cite{Yugay2023Gaussian-SLAM} {,}
             Huang et al. \cite{Huang2023}
             ,level5_2]
        ]
        [Mesh Extraction\\Physics (Sec. \ref{tree:5}), level2
            [Xie et al. \cite{Xie2023} {,}
             Guadon et al. \cite{guadon2023sugar} {,}
             Duisterhof et al. \cite{Duisterhof2023} {,}
             Chen et al. \cite{Chen2023}
             ,level5_2]
        ]
        [Editable\\(Sec. \ref{tree:7}), level2
            [Chen et al. \cite{Chen2023_Swift} {,}
             Fang et al. \cite{Fang2023} {,}
             Huang et al. \cite{Huang2023} {,}
             Ye et al. \cite{Ye2023} {,}\\
             Cen et al. \cite{Huang2023} {,}
             Zhou et al. \cite{Zhou2023}
             ,level5_2]
        ]
    ]
]
\end{forest}
\caption{Taxonomy of selected key Gaussian Splatting innovation papers, selected using a combination of citations and GitHub star rating.}
\label{fig:taxonomy_innovations}
\end{figure*}
\section{Functional Advancements}
\label{sec:fun_adv}
This section examines the advances in functional capabilities that have been achieved since the Gaussian Splatting algorithm was first introduced.
\subsection{Dynamic and deformation}
\label{tree:1}
\begin{figure*} 
\centering
\subfloat[Persistent Dynamic NVS and Tracking Results from~\cite{Luiten2023}.]{\includegraphics[width=0.8\textwidth]{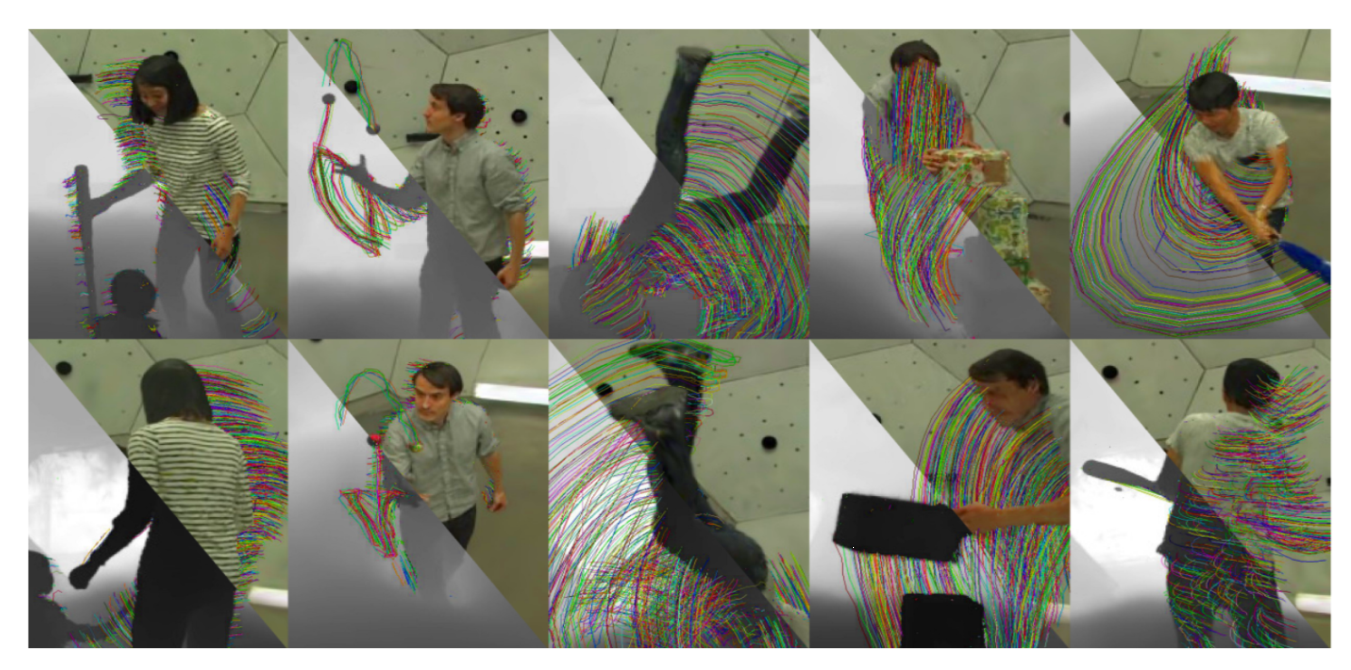}} \\
\subfloat[Pipeline of Control4D: utilizing GaussianPlanes~\cite{Shao2023}.]{\includegraphics[width=0.8\textwidth]{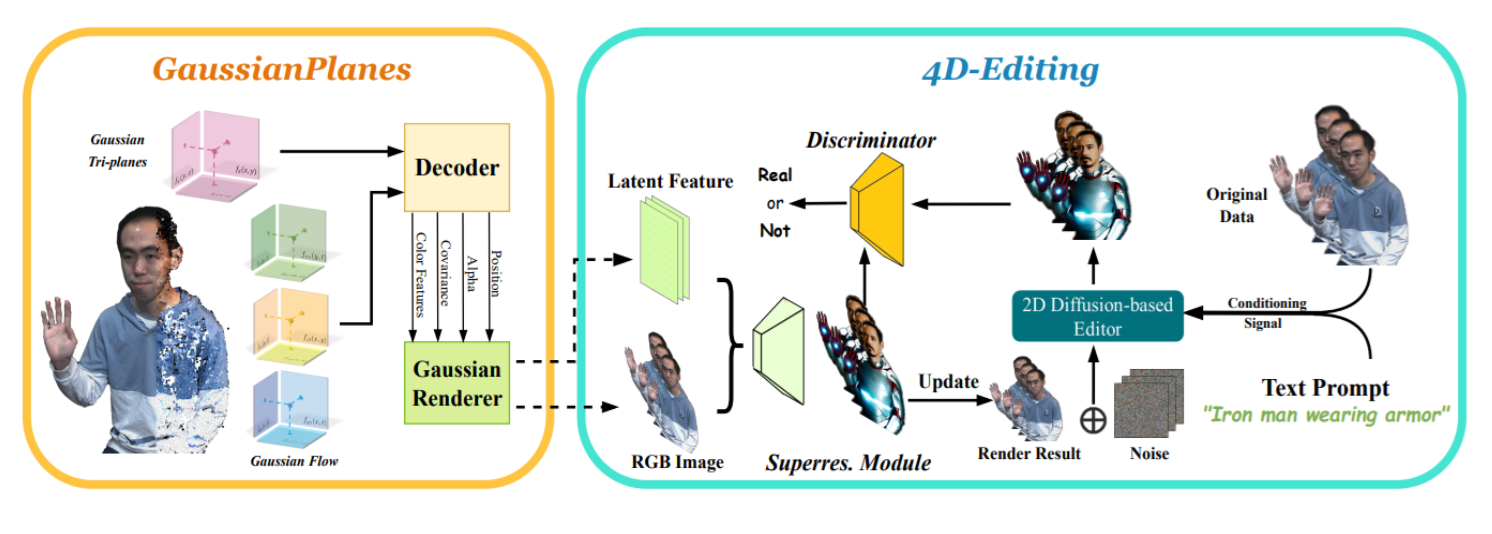}} \\
\subfloat[Proposed method by Yang et al. can reconstruct accurate dynamic scene geometry and render high-quality images in both the NVS, and time interpolation compared with HyperNeRF~\cite{yang2023deformable3dgs}.]{\includegraphics[width=0.8\textwidth]{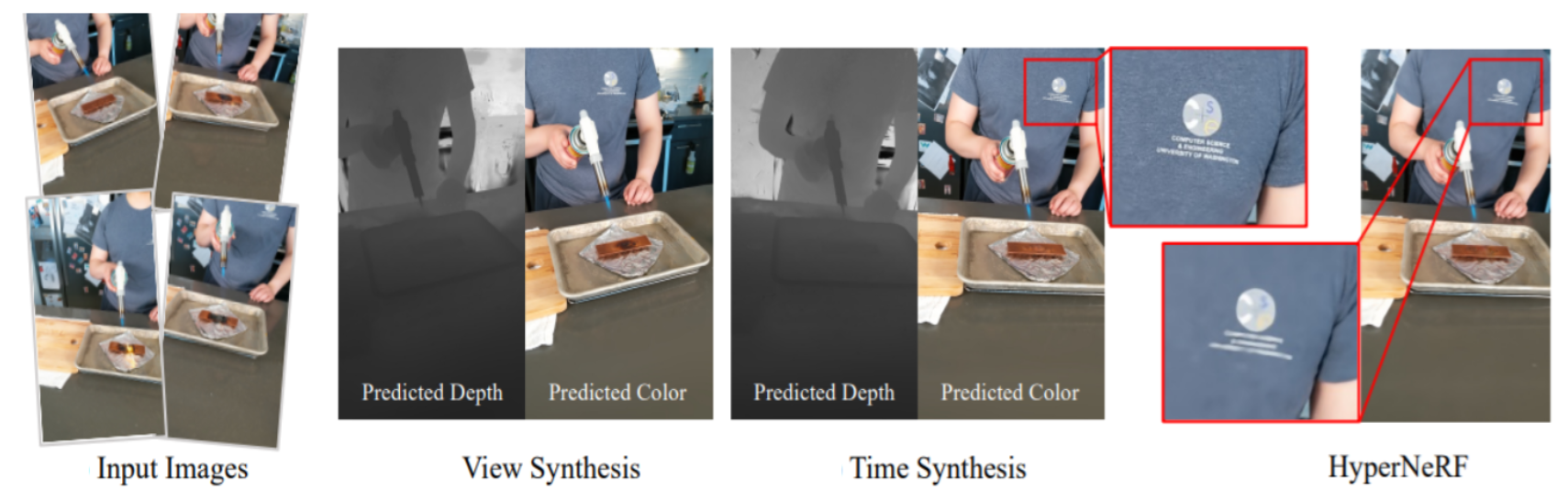}}
\caption{\textbf{Dynamic and deformation based methods.}}
\label{fig:dynamic}
\end{figure*}

In contrast to general Gaussian splats, where all parameters of the 3D covariance matrix are dependent on only the input images, in this case, to capture the dynamic of the splats over time, some of the parameters are dependent on time or time step~\cite{Katsumata2023}. For example, the position is time-step or frame dependent. This position can be updated by the next frame in a temporally consistent manner. Also some latent encoding can be learnt which can be used to edit or propagate the Gaussian in each time-step during render to achieve certain effect like expression changes in a avatar~\cite{Shao2023, Xiong2023_SparseGS}, and application of force to a non rigid body~\cite{yang2023deformable3dgs, Das2023}. Figure \ref{fig:dynamic} shows a few of the dynamic and deformation-based methods.

The dynamic and deformable models can be easily represented by a slight modification to the original Gaussian Splatting representation:\\

\begin{minipage}{\columnwidth}
\centering
\vspace{0.5mm}
\raggedright
1) A 3D position at time $t$: $[x(t), y(t), z(t)]^{\sf T} \in \mathbb{R}^{3}$, \\
2) A 3D rotation at time $t$ represented by a quaternion: $[q_x(t), q_y(t), q_z(t), q_w(t)]^{\sf T} \in \mathbb{R}^{4}$ \\ 
3) A scaling factor: $[s_x, s_y, s_z]^{\sf T} \in \mathbb{R}^{3}$\\
4) Spherical harmonics coefficients representing color with the degrees of freedom $k$: $h \in \mathbb{R}^{3 \times (k+1)^2}$ \\
5) An opacity: $o \in \mathbb{R}$
\vspace{0.5mm}
\end{minipage}
\subsubsection{Motion and Tracking}
\label{tree:1_1}
Most of the work related to dynamic Gaussian Splatting extends on the motion tracking of the 3D gaussian across time steps, instead of having a separate splat for each time step. Katsumata et al. proposed  a \textbf{Fourier approximation for the position and linear approximation for the rotation quaternion} in~\cite{Katsumata2023}.

The paper by Luiten et al.~\cite{Luiten2023} introduces a method for \textbf{capturing full 6 degrees of freedom} for all 3D points in dynamic scenes. By incorporating local-rigidity constraints, Dynamic 3D Gaussians represent consistent space rotation, enabling dense \textbf{6-DOF tracking} and reconstruction without correspondence or flow input. The method outperforms PIPs in 2D tracking~\cite{harley2022particle}, achieving a 10x lower median trajectory error, higher trajectory accuracy, and a 100\% survival rate. This versatile representation facilitates applications like 4 dimensional video editing, first person view synthesis, and dynamic scene generation.

Wu et al. in~\cite{Wu2023} proposes a novel approach known as 4D Gaussian Splatting (4D-GS). The author proposes a \textbf{Spatial-Temporal Structure Encoder and a Multi-head Gaussian Deformation Decoder.} This holistic representation combines both 3D Gaussians and 4D neural voxels, enabling real-time rendering at high resolutions. The method achieves a notable frame rate of 82 frames per seconds (FPS) at a resolution of  800$\times$800 using an RTX 3090 GPU, maintaining good quality. Despite its success in rapid convergence and real-time rendering, 4D-GS faces challenges in optimizing Gaussians for large motions, dealing with the absence of background points, and addressing imprecise camera poses. Additionally, the method struggles to separate the joint motion of static and dynamic Gaussian parts under monocular settings without additional supervision. Lastly, there is a need for a more compact algorithm to handle urban-scale reconstruction due to the substantial querying of Gaussian deformation fields by a large number of 3D Gaussians.

To properly represent spatial and temporal structures in dynamic scenes Yang et al. in~\cite{yang2023deformable3dgs} propose a holistic approach, treating spacetime as a whole. They advocate \textbf{approximating the underlying spatiotemporal 4D volume of dynamic by optimizing a set of 4D primitives, incorporating explicit geometry and appearance modeling.} The proposed model is conceptually straightforward, utilizing a 4D Gaussian parameterized by anisotropic ellipsoids capable of arbitrary rotation in space and time. Additionally, it incorporates view-dependent and time-evolved appearance represented by the coefficients of 4D spherical harmonics. This approach offers simplicity, flexibility for variable-length videos, end-to-end training, and efficient real-time rendering, making it well-suited for capturing complex dynamic scene motions. 

Kratimenos et al. in~\cite{Kratimenos2023} effectively addresses the challenge of the motion field in dynamic scenarios, which is naturally underconstrained, guaranteeing effective optimization. In order to do this, \textbf{each point is bound to motion coefficients that enforce the sharing of basis trajectories.} The \textbf{introduction of a sparsity loss to the motion coefficients enables the disentanglement of scene motions, providing independent control and the generation of novel motion combinations}. Remarkably, state-of-the-art render quality is achieved in minutes of training, and the model can synthesize high-quality views of dynamic scenes with superior photorealistic results when trained less than thirty minutes. Their proposed representation is characterized by interpretability, efficiency, and expressiveness, allowing for real-time NVS including dynamic motions in scene in both monocular and multi-view scenarios.

Lin et al. introduce a novel \textbf{Dual-Domain Deformation Model (DDDM)} in~\cite{Lin2023} which is explicitly designed to model attribute deformations for each Gaussian point. This model uses Fourier series fitting in the frequency domain and polynomial fitting in the time domain to capture time-dependent residuals. \textbf{The DDDM is adept at handling deformations across complex video scene, eliminating the need to train individual 3D Gaussian Splatting (3D-GS) models for every frame.} Notably, discretized Gaussian point explicit deformation modeling guarantees quick training and 4D scene rendering, similar to the original 3D-GS intended for static 3D reconstruction. This approach have  substantial efficiency improvement, with almost a $5\times$ faster training speed compared to 3D-GS modeling. However, there is an identified opportunity for enhancement in maintaining high-fidelity thin structures in the final rendering.

\subsubsection{Expression or Emotion variation and Editable in Avatars}
\label{tree:1_2}
Shao et al. introduce GaussianPlanes in~\cite{Shao2023}, a 4D representation through plane-based decomposition in both 3D space and time, improving effectiveness in 4D editing. Additionally, Control4D leverages a \textbf{4D generator to optimize the continuous creation space from inconsistent photos,} resulting in better consistency and quality. The proposed method employs \textbf{GaussianPlanes to train the implicit representation of a 4D portrait scene,} followed by rendering into latent features and RGB images using Gaussian rendering. A generative adversarial network (GAN)~\cite{goodfellow2020generative} based generator, along with a 2D-diffusion-based editor, refines the dataset and produces real and fake images for discrimination. The discriminative results contribute to iterative updates of both the Generator and Discriminator. However, the approach faces challenges in handling rapid and extensive non-rigid movements due to relying on canonical Gaussian point clouds with flow representation. The method is constrained by ControlNet~\cite{zhang2023adding}, limiting edits to a coarse level and preventing precise expression or action edits. Furthermore, the editing process requires iterative optimizations, lacking a single-step solution.

Huang et al. utilizes \textbf{sparse control points, a fraction of the Gaussians, to learn compact 6 DoF transformation bases} in~\cite{Huang-sc-gs}. These bases, locally interpolated with learned weights, define the motion field of 3D Gaussians. A deformation MLP predicts time-varying 6 DoF transformations for each control point, simplifying learning, improving capabilities, and ensuring coherent motion patterns. The joint learning process encompasses 3D Gaussians, canonical space locations of control points, and the deformation MLP, reconstructing appearance, geometry, and dynamics. Adaptive adjustments to control point locations and numbers accommodate motion complexities, \textbf{with an As-Rigid-As Possible Regularization (ARAP) loss enforcing spatial continuity and local rigidity.} The explicit sparse motion representation allows user-controlled motion editing while maintaining high-fidelity appearances. Experimental results showcase superiority in NVS with high rendering speed and novel appearance-preserved motion editing applications. However, the method's performance is susceptible to inaccurate camera poses, leading to reconstruction failures. Additionally, the method's testing has been limited to scenes with modest motion changes, and extending its applicability to intense movements remains an area for exploration.

Yu et al. in~\cite{Yu2023} introduces a \textbf{Controllable Gaussian Splatting method (CoGS)}, providing real-time handling of elements in a scene without the need for pre-computed control signals.
\subsubsection{Non-Rigid or deformable objects}
\label{tree:1_3}
Implicit neural representation has brought around a significant transformation in dynamic scene reconstruction and rendering. Nevertheless, contemporary methods in dynamic neural rendering encounter challenges related to capturing intricate details and achieving real-time rendering in dynamic scenes.

In response to these challenges, Yang et al. proposed \textbf{Deformable 3D Gaussians for High-Fidelity Monocular Dynamic Scene Reconstruction} in~\cite{yang2023deformable3dgs}. A novel deformable 3D-GS method is proposed. This method \textbf{utilizes 3D Gaussians learned in canonical space with a deformation field, specifically designed for monocular dynamic scenes.} The approach introduces an annealing smooth training (AST) mechanism tailored for real-world monocular dynamic scenes, effectively addressing the effect of erroneous poses on time interpolation tasks without introducing additional training overhead. By using a differential Gaussian rasterizer, the deformable 3D Gaussians not only enhance rendering quality but also achieve real-time speeds, surpassing existing methods in both aspects. The method proves to be well-suited for tasks such as NVS and offers versatility for post-production tasks due to its point-based nature. The experimental results underscore the method's superior rendering effects and real-time capabilities, confirming its efficacy in dynamic scene modeling.

Das et al. in~\cite{Das2023} introduces \textbf{NPGs (Neural Parametric Gaussians)}, that address the challenging task of reconstructing dynamic objects from monocular videos. The approach involves a two-stage process: \textbf{first, fitting a low-rank neural deformation model to preserve consistency in NVS, and second, optimizing 3D Gaussians driven by the coarse model for high-quality reconstruction.} Their model is based on a local representation of temporally shared anchored 3D Gaussian where the local oriented volumes causes the deformation. The resulting radiance fields enable photo-realistic high-quality reconstructions of non-rigidly deforming objects, ensuring uniformity across NVS. NPGs exhibit superior performance, specially in scenes with limited multi-view cues.

\subsection{Diffusion}
\label{tree:2}
Diffusion and Gaussian Splatting is a powerful technique for generating 3D objects from text descriptions/prompts. It combines the strengths of two different approaches: diffusion models and Gaussian Splatting. Diffusion models are a type of neural network that can learn to generate images from a noisy input \cite{ho2020denoising}. By feeding the model a sequence of increasingly clean images, the model learns to reverse the process of image corruption, eventually generating a clean image from a completely random input. This can be used to generate images from text descriptions, as the model can learn to associate words with the corresponding visual features. The text-to-3D with diffusion and Gaussian Splatting pipeline works by first using a diffusion model to generate an initial 3D point cloud from the text description. The point cloud is then converted into a set of Gaussian spheres using Gaussian Splatting. Finally, the Gaussian spheres are rendered to produce a 3D image of the object.

\subsubsection{Text based generation}
\label{tree:2_1}
The work by Yi et al. introduces \textbf{Gaussian-Dreamer} in~\cite{Yi2023}, a text-to-3D method that seamlessly connects 3D and 2D diffusion models through Gaussian splitting, ensuring both 3D consistency and intricate detail generation. Figure \ref{fig:dream_gaussian} shows the proposed model generating images. \textbf{To further enrich content, noisy point growing and color perturbation are introduced to supplement the initialized 3D Gaussians.} The method is characterized by its simplicity and effectiveness, generating a 3D instance within 15 minutes on a single GPU, showcasing superior speed compared to previous methods. The generated 3D instance can be directly rendered in real time, highlighting the practical applicability of the approach. The overall framework involves initialization with 3D diffusion model priors and optimization with the 2D diffusion model, enabling the creation of high-quality and diverse 3D assets from text prompts by leveraging the strengths of both diffusion models.
\Figure[hbt!](topskip=0pt, botskip=0pt, midskip=0pt)[width=0.99\columnwidth]{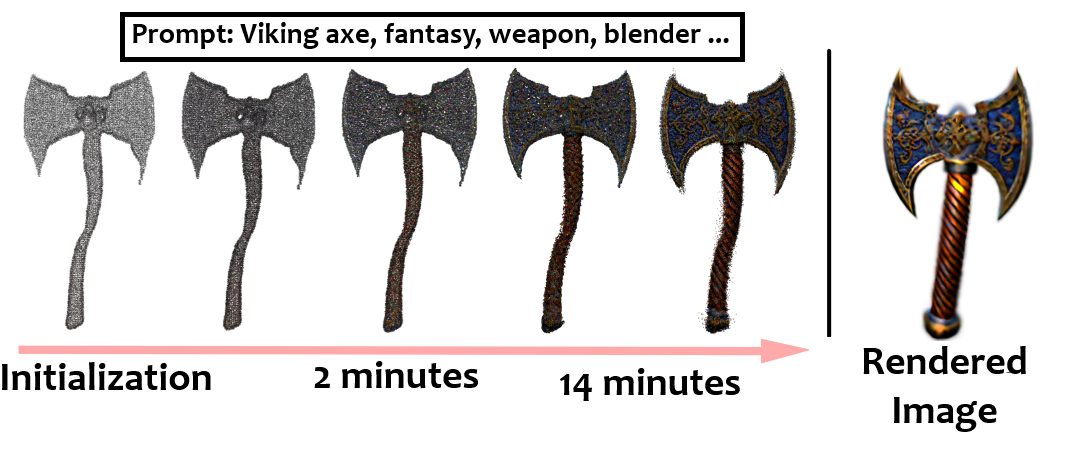}
{\textbf{Dream-Gaussian framework generating images through iteration~\cite{Yi2023}.}\label{fig:dream_gaussian}}

Chen et al. presented Gaussian Splatting based text-to-3D GENeration \textbf{(GSGEN)} in~\cite{Chen2023_text}, a text-to-3D generation method utilizing 3D Gaussians as a representation. By leveraging \textbf{geometric priors, emphasizing the unique advantages of Gaussian Splatting} in text-to-3D generation. \textbf{The two-stage optimization strategy combines joint guidance of 2D and 3D diffusion for shaping a coherent rough structure in geometry optimization, followed by densification in appearance refinement based on compactness.} GSGEN is validated across various textual prompts, demonstrating its ability to generate 3D assets with more accurate geometry and enhanced fidelity. Notably, GSGEN excels in capturing high-frequency components in objects, such as feathers, intricate textures, and animal fur. However, challenges arise when the provided text prompt is complex or involves complicated logic, given the limited language understanding of Point-E~\cite{nichol2022point} and the CLIP~\cite{radford2021learning} text encoder used in Stable Diffusion. Although incorporating 3D priors mitigates the Janus problem\footnote{The Janus dilemma pertains to the difficulty of addressing temporal inconsistencies or uncertainties in data, especially in situations where historical records or information may lack completeness or contain contradictions. This poses a notable hindrance in areas like historical research, data analysis, and artificial intelligence, necessitating the use of specialized methodologies to effectively reconcile conflicting temporal data points.}, potential degenerations persist, particularly with extremely biased textual prompts in the guidance diffusion models.

Tang et al. introduce a pioneering framework~\cite{Tang2023} for 3D content creation by integrating Gaussian Splatting into generative settings, leading to a significant reduction in generation time compared to optimization-based 2D lifting methods. Additionally, the authors present an efficient \textbf{mesh extraction algorithm from 3D Gaussians and a UV-space texture refinement} stage to further elevate the quality of the generated content. Through extensive experiments encompassing both image-to-3D and text-to-3D tasks, this method demonstrates a remarkable balance between optimization time and generation fidelity, opening up new possibilities for real-world deployment of 3D content generation. It's important to note that, like previous text-to-3D approaches, the authors encounter common challenges such as the Multi-face Janus problem and issues related to baked lighting.

Liang et al. proposed an analysis of \textbf{Score Distillation Sampling (SDS)} in text-to-3D generation~\cite{Liang2023}, exposing its limitations. They introduce Interval Score Matching (ISM) to outperform SDS, integrating it with 3D Gaussian Splatting for state-of-the-art performance in various applications, achieving realism with reduced training costs.

Chung et al. introduces \textbf{LucidDreamer} in~\cite{Chung2023_domain} shown in Figure \ref{fig:lucid_dreamer}, a domain-free 3D scene generation framework. Leveraging Stable Diffusion, depth estimation, and explicit 3D representation, LucidDreamer demonstrates superior domain generalization with diverse input types, generating high-quality 3D scenes across scenarios.
\Figure[hbt!](topskip=0pt, botskip=0pt, midskip=0pt)[width=0.99\columnwidth]{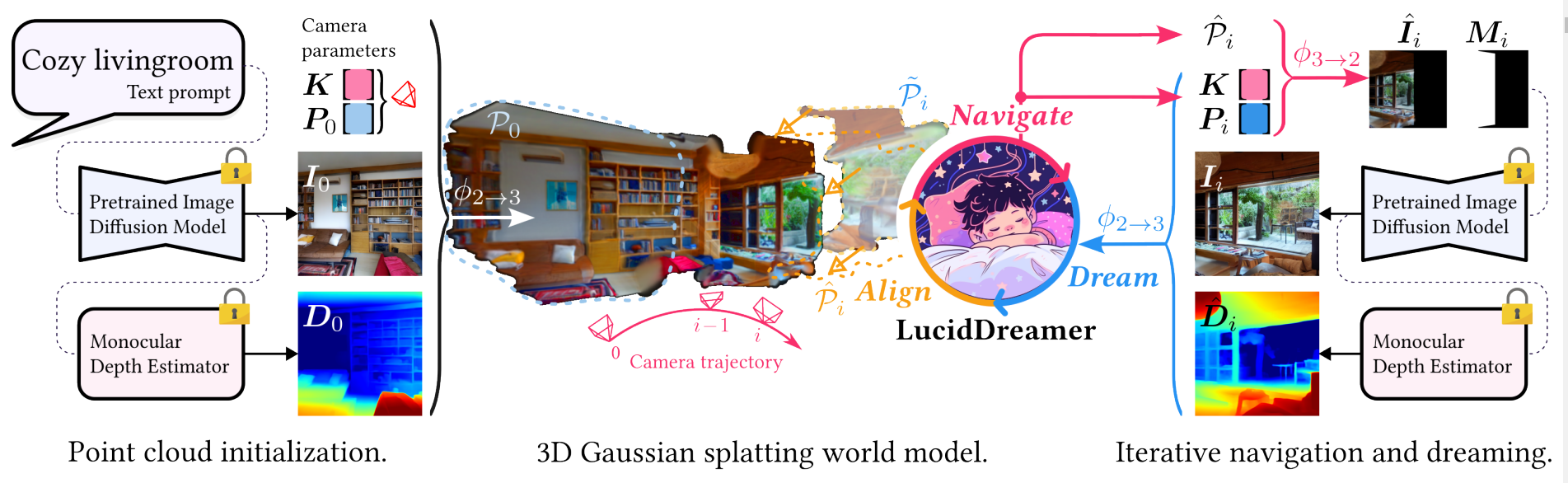}
{\textbf{Lucid Dreamer framework with iterative navigation and dreaming~\cite{Chung2023_domain}.}\label{fig:lucid_dreamer}}

The \textbf{HumanGaussian Framework} in~\cite{Moreau2023} by Moreau et al. generates 3D humans from text prompts using the neural representation of 3D-GS. The Structure-Aware SDS optimizes appearance and geometry, achieving efficient and effective 3D human generation with fine-grained geometry and realistic appearance.

The proposed \textbf{Compositional generation for text-to-3d via gaussian splatting (CG3D)} framework~\cite{Vilesov2023} by Vilesov et al. introduces a text-driven compositional 3D scene generation method emphasizing scalability and physical realism. Leveraging explicit radiance fields, CG3D achieves coherent multi-object scenes, allowing quick editing through text prompts while addressing challenges like the Janus problem. Future work aims to enhance the support for intricate object interactions and large-scale compositions.
\subsubsection{Denoising and Optimisation}
\label{tree:2_2}
The \textbf{GaussianDiffusion} framework in~\cite{Li2023} by Li et al. represents a novel text-to-3D approach, leveraging Gaussian Splatting and Langevin dynamics diffusion models to accelerate rendering and achieve unparalleled realism. The introduction of structured noise addresses multi-view geometric challenges, while the variational Gaussian Splatting model mitigates convergence issues and artifacts. While current results show improved realism, ongoing research aims to refine aspects of blurriness and haze introduced by variational Gaussian for further enhancement.

Yang et al. encompass a thorough examination of existing diffusion priors, leading to the proposition of~\cite{Yang2023prior}, an \textbf{unified framework that improves these priors by optimizing denoising scores.} The versatility of the approach extends across various use cases, consistently delivering substantial performance enhancements. In experimental evaluations, our method achieves unprecedented performance on ~\cite{he2023t}, surpassing contemporary methods. Despite its success in refining the texture aspect of 3D generation, there is room for improvement in enhancing the geometry of the generated 3D models.
\subsection{Optimization and speed}
\label{tree:6}
This subsection will deal with the techniques researchers have developed for faster training and/or inference speed. In the study by Chung et al.~\cite{Chung2023}, a method is introduced to optimize Gaussian Splatting for 3D scene representation using a \textbf{limited number of images while mitigating overfitting issues}. The conventional approach of representing a 3D scene with Gaussian splats can lead to overfitting, particularly when the available images are limited. This technique~\cite{Chung2023} uses the depth map from  a pre-trained monocular depth estimation model as a geometric guide, and align with sparse feature points from a SFM pipeline. These helps in optimization of 3D Gaussian Splatting, reducing floating artifacts and ensuring geometric coherence. The proposed depth-guided optimization strategy is tested on the LLFF~\cite{mildenhall2019llff} dataset, showcasing improved geometry compared to if only images are used. The study includes the introduction of an \textbf{early stop strategy and a smoothness term for the depth map, both contributing to enhanced performance.} However, limitations are acknowledged, such as reliance on the accuracy of the monocular depth estimation model and the dependency on COLMAP's~\cite{COLMAP} performance. Future work is suggested to explore interdependent estimated depths and address challenges in areas with difficult depth estimation, like textureless plains or the sky.

The study by Lee et al. introduces a compact 3D Gaussian representation framework~\cite{Lee2023}, leveraging the advantages of 3D-GS. While 3D-GS offers rapid rendering and promising image quality, it demands a significant number of 3D Gaussians, leading to substantial memory and storage requirements. The proposed framework \textbf{uses a learnable mask strategy that considerably reduces the number of Gaussians without sacrificing performance.} Additionally, a grid-based neural field is introduced for a compact representation of view-dependent color, and codebooks are learned to compress geometric attributes effectively. The experiments demonstrate over a 10× reduction in storage, enhanced rendering speed, and maintained scene representation quality compared to 3D-GS. The framework's emphasis on reducing Gaussian points and compressing attributes establishes it as a comprehensive solution, fostering broader adoption in fields requiring efficient and high-quality 3D scene representation.

In their research~\cite{Girish2023}, Girish et al. introduced a novel technique utilizing quantized embedding for efficient memory utilization. Girish et al. used a \textbf{coarse-to-fine strategy for optimized Gaussian point clouds}, achieving scene representations with fewer Gaussians and quantized attributes, resulting in faster training and rendering speeds. Validating across various datasets and scenes with a notable 10-20× reduction in memory usage and improved training/inference speed. Contributions include a novel compression method for 3D Gaussian point clouds, optimization enhancements via opacity coefficient quantization, progressive training, and controlled densification. Ablation studies underscore the effectiveness of these components, with the approach demonstrating comparable quality to 3D-GS while being faster and more efficient. Overall, this method represents a significant advancement in 3D reconstruction and NVS, striking a balance between efficiency and reconstruction quality.

In the study~\cite{Fu2023} Fu et al. introduces \textbf{COLMAP-Free 3D Gaussian Splatting (CF-3DGS)}, a novel end-to-end framework for simultaneous camera pose estimation and NVS from sequential images, addressing challenges posed by large camera motions and lengthy training durations in previous approaches. Diverging from the implicit representations of NeRFs, \textbf{CF-3DGS leverages explicit point clouds to represent scenes. The method sequentially processes input frames, progressively expanding 3D Gaussians to reconstruct the entire scene, demonstrating enhanced performance and robustness on challenging scenes, such as 360° videos.} The approach optimizes camera pose and 3D-GS jointly in a sequential manner, making it particularly suitable for video streams or ordered image collections. The utilization of Gaussian Splatting enables rapid training and inference speeds, showcasing the advantages of this approach over previous methods. While demonstrating effectiveness, it's acknowledged that the sequential optimization restricts the application primarily to ordered image collections, leaving room for exploration into extensions for unordered image collections in future research.

\subsection{Rendering and shading methods}
\label{tree:8}
Yu et al. in~\cite{Yu2023Mip_Splatting} observed in 3D-GS, specifically artifacts in NVS results when changing the sampling rate. The introduced solution involves incorporating a \textbf{3D smoothing filter to regulate the maximum frequency of 3D Gaussian primitives}, resolving artifacts in out-of-distribution renderings. Additionally, the 2D dilation filter is replaced with a 2D Mip filter to address aliasing and dilation issues. Evaluations on benchmark datasets demonstrate the effectiveness of Mip-Splatting, particularly when modifying the sampling rate. The proposed modifications are principled and straightforward, requiring minimal changes to the original 3D-GS code. However, there are acknowledged limitations, such as errors introduced by the Gaussian filter approximation and a slight increase in training overhead. The research presents Mip-Splatting as a competitive solution, demonstrating its performance parity with state-of-the-art methods and superior generalization in out-of-distribution scenarios, showcasing its potential in achieving alias-free rendering at arbitrary scales.

Gao et al.  proposes a novel approach in~\cite{Gao2023} to \textbf{3D point cloud rendering that enables material and lighting decomposition from multi-view images.} This framework supports editing, ray tracing, and real-time relighting of the scene in a differentiable manner. Each point in the scene is represented by a "relightable" 3D Gaussian, carrying information about its normal direction, material properties like bidirectional reflectance distribution function(BRDF), and incoming light from various directions. For accurate lighting estimation, the incoming light is separated into global and local components, considering visibility based on the viewing angle. Scene optimization leverages 3D Gaussian Splatting, while physically-based differentiable rendering handles BRDF and lighting decomposition. An innovative point-based ray-tracing method utilizing a bounding volume hierarchy enables efficient visibility baking and realistic shadows during real-time rendering. Experiments demonstrate superior BRDF estimation and novel view rendering compared to existing methods. However, challenges remain for scenes without clear boundaries and the requirement of object masks during optimization. Future work could explore integrating multi-view stereo (MVS) cues to improve the geometric accuracy of point clouds generated by 3D Gaussian Splatting. This "Relightable 3D Gaussian" pipeline demonstrates promising real-time rendering capabilities and opens doors for revolutionizing mesh-based graphics with a point cloud-based approach that allows for relighting, editing, and ray tracing.

Liang et al. proposes \textbf{3D Gaussian Splatting for Inverse Rendering (GS-IR)} in~\cite{Liang2023GS_IR},  a novel inverse rendering approach leveraging the strengths of 3D-GS, a powerful technique for generating novel views. Unlike approaches relying on implicit neural representations and volume rendering, GS-IR expands the capabilities of 3D-GS to directly estimate scene geometry, material properties, and lighting conditions from multi-view images, even under unknown lighting. \textbf{It successfully addresses challenges like normal estimation and occlusion handling through an efficient optimization scheme that combines depth-based regularization and baking-based occlusion modeling.} The inherent flexibility of 3D-GS enables fast and compact reconstruction of the scene geometry, leading to photorealistic NVS and physically accurate rendering. Extensive evaluations across various scenes demonstrate that GS-IR outperforms existing methods, achieving state-of-the-art results in both reconstruction quality and efficiency.

Yan et al. proposed a \textbf{multi-scale 3D Gaussian Splatting algorithm to address the degradation in rendering quality and speed} that occurs at lower resolutions or from a faraway camera position in~\cite{Yan2023Multi_Scale}. Acknowledging the aliasing effect caused by pixel size falling below the Nyquist frequency, the algorithm maintains Gaussians at different scales to represent the scene effectively. Inspired by mipmap and  levels of detail (LOD) algorithms, larger, coarser Gaussians are added for lower resolutions by aggregating smaller and finer Gaussians from higher resolutions. This approach achieves significant improvements compared to standard 3D Gaussian Splatting. It shows a 13\%-66\% boost in PSNR and a 160\%-2400\% increase in rendering speed across resolutions ranging from 4x to 128x. While there is some initial overhead during splatting, the method effectively reduces aliasing artifacts and significantly improves rendering efficiency. Future research will explore lightweight filtering criteria for Gaussian functions to further enhance speed. Overall, this algorithm demonstrates effectiveness in both rendering quality and speed at various resolutions, overcoming limitations of previous 3D Gaussian Splatting methods.

Jiang et al. introduced \textbf{GaussianShader} in ~\cite{Jiang2023_3d_gaussian} which integrates a simplified shading function directly onto 3D Gaussians, improving the visual quality of rendered reflective scenes while maintaining efficiency in both training and rendering. \textbf{Estimating accurate normals on discrete 3D Gaussians has been challenging, this approach overcomes this difficulty with a novel framework leveraging shortest axis directions and a customized loss function for consistency.} GaussianShader achieves a remarkable balance between rendering quality and efficiency, outperforming standard 3D Gaussian Splatting in terms of PSNR on datasets containing specular objects. Additionally, it demonstrates significant optimization time improvements compared to previous methods. By explicitly approximating the rendering equation, GaussianShader enhances realism and allows for real-time rendering, making it suitable for interactive applications. In summary, this method is a significant step forward in rendering 3D objects, particularly for reflective surfaces, by combining shading functions with an extended 3D Gaussian model and introducing an innovative normal prediction technique for high-quality results.

Lu et al. introduced \textbf{Scaffold-GS} in~\cite{Lu2023}, for rendering complex scenes effectively. Their method utilizes anchor points to strategically distribute local 3D Gaussians and dynamically predicts their attributes based on viewing conditions. Through a clever "growing and pruning" strategy for these anchors, Scaffold-GS efficiently adapts its representation to the scene, minimizing redundant Gaussians. This results in improved rendering quality while handling scenes with varying details and viewpoints without compromising speed.

3D Gaussian Splatting faces challenges in few-shot scenarios, where it tends to overfit to training views, leading to issues like background collapse and excessive floaters. In response, Xiong et al. introduced~\cite{Xiong2023_SparseGS} proposed to \textbf{enable the coherent training of 3D-GS based radiance fields for 360-degree scenes using sparse training views.} The method integrates depth priors with generative and explicit constraints to address challenges such as background collapse and floater artifacts, enhancing consistency from unseen viewpoints. Experimental results demonstrate the superiority of the proposed technique over base 3D-GS and NeRF-based methods in terms of LPIPS on the MipNeRF-360 dataset, achieving substantial improvements with reduced training and inference costs. Despite its reliance on the initial point cloud from COLMAP, the method showcases impressive performance in few-shot NVS, with potential for further improvements through investigations into point cloud densification techniques.

In response to the persistent challenge of efficient NVS from limited observations, Zhu et al. propose a Few-Shot View Synthesis framework in~\cite{Zhu2023}. To achieve real-time and photorealistic results with only three training views. This adeptly handles the sparsity of \textbf{initialized SFM points through a well-designed Gaussian Unpooling process}, iteratively distributing new Gaussians around representative locations to fill in local details in vacant areas. The framework incorporates a large-scale pretrained monocular depth estimator within the Gaussian optimization process, using online augmented views to guide geometric optimization for an optimal solution. The accuracy and rendering efficiency tests are performed on various datasets, including LLFF~\cite{mildenhall2019llff}, Mip-NeRF360~\cite{barron2022mipnerf360}, and custom dataset generated using Blender. Noteworthy features include a novel point-based framework with Proximity-guided Gaussian Unpooling for comprehensive scene coverage, integration of monocular depth priors for optimized Gaussian representation, and real-time rendering speed (200+ FPS) with improved visual quality. This framework paves the way for practical applications in real-world scenarios, offering a valuable contribution to the field of few-shot view synthesis.

\subsection{Compression}
\label{tree:9}
Fan et al. in~\cite{Fan2023} introduce a novel technique for compressing 3D Gaussian representations used in rendering. Their method identifies and removes redundant Gaussians on the basis of their significance, similar to network pruning, ensuring minimal impact on visual quality. Leveraging knowledge distillation and pseudo-view augmentation, \textbf{LightGaussian transfers information to a lower-complexity representation with fewer spherical harmonics, further reducing redundancy. Additionally, a hybrid scheme called VecTree Quantization optimizes the representation by quantizing attribute values, leading to even smaller size without significant loss in accuracy.} Compared to standard approaches, LightGaussian achieves an average compression ratio of over 15x, significantly boosting rendering speed from 139 FPS to 215 FPS on datasets like Mip-NeRF 360~\cite{barron2022mipnerf360}, and Tanks\&Temples~\cite{Knapitsch2017}. The key steps involved are calculating global significance, pruning Gaussians, distilling knowledge with pseudo-views, and quantizing attributes using VecTree. Overall, LightGaussian offers a groundbreaking solution for converting large point-based representations into a compact format, resulting in dramatic reductions in data redundancy and substantial improvements in rendering efficiency.

Navaneet et al. propose a simple yet effective solution in~\cite{Navaneet2023}, \textbf{leveraging vector quantization based on the K-means algorithm to quantize Gaussian parameters.}
The approach involves storing a small codebook alongside the index of the code for each Gaussian, followed by further compression of indices through sorting and a method akin to run-length encoding.
Through extensive experiments on standard and a larger-than-standard benchmark, the method demonstrates its effectiveness in reducing the storage cost of the original 3D Gaussian Splatting by nearly 20×, with minimal impact on the quality of rendered images.
This compression technique provides a valuable trade-off, maintaining the efficiency of 3D Gaussian Splatting while significantly mitigating storage demands.
\section{Applications and Case Studies}
\label{sec:app_adv}
This section delves into the notable advancements in applications of Gaussian Splatting since the inception of the algorithm in July 2023. These advancements have found specific utility in various domains, such as avatars, SLAM, and mesh extraction and physics simulation. Gaussian Splatting, when applied to these specialized use cases, demonstrates its versatility and effectiveness in diverse application scenarios.
\subsection{Avatars}
\label{tree:3}
A large amount of research in Gaussian Splatting is focused towards developing digital avatars of human, pertaining to the rise of AR/VR application boom. Capturing a subject from less number of viewpoint and constructing a 3D model is a challenging task and Gaussian Splatting is helping researcher and industries to achieve that.
\subsubsection{Joint angles or articulation}
\label{tree:3_1}
These kind of Gaussian Splatting technique focuses on modeling the human body in terms of joint angles. Some of the parameter of these kind of model reflects the 3D joint position, the angles, and other similar parameters. The input frame is decoded to find out the current frame 3D joint position and angle.\\
Zielonka et al. presents a model in~\cite{Zielonka2023} for human body representation using Gaussian splats, achieving real-time rendering with the innovative 3D-GS technique. Unlike existing photorealistic drivable avatars, \textbf{Drivable 3D Gaussian Splatting (D3GA) doesn't rely on accurate 3D registrations during training or dense input images during testing. Instead, it leverages dense calibrated multi-view videos for real-time rendering and introduces tetrahedral cage-based deformations driven by keypoints and angles in joint, making effective for applications involving communication as shown in Figure \ref{fig:d3ga}.} The experiment includes subjects with various clothing, body shapes, and motions, where D3GA outperforms other state-of-the-art methods, showcasing superior pose-based avatar generation for dense multi-view scenes without the need for ground truth registration. The contributions include the first implementation of DG3A, tetrahedral cage-based deformations, and state-of-the-art pose-based avatar generation without ground truth registration. D3GA demonstrates high-quality results and promising advancements in geometry and appearance modeling for dynamic sequences without relying on ground truth geometry, thereby streamlining the data processing pipeline.
\Figure[hbt!](topskip=0pt, botskip=0pt, midskip=0pt)[width=0.99\columnwidth]{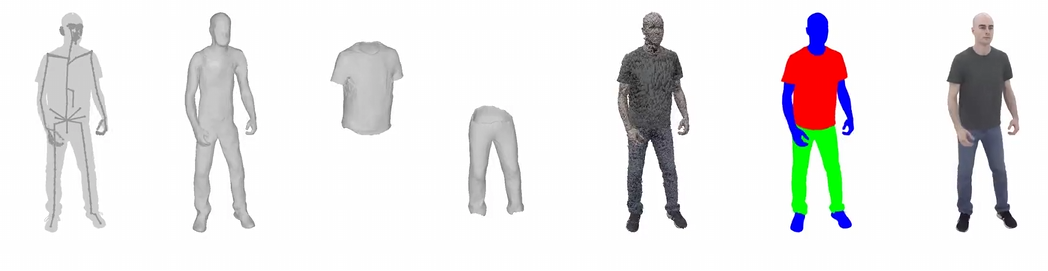}
{\textbf{D3GA framework, from the left: joint angles, predicted body cage, pred upper cage, pred lower cage, 3D Gaussians, garment parts, final image~\cite{Zielonka2023}.}\label{fig:d3ga}}

Jena et al. in~\cite{Jena2023} extended the skinning of the underlying \textbf{Skinned Multi-Person Linear (SMPL) geometry to arbitrary locations} in the canonical space to model the human articulation.
Lei et al. similarly in \textbf{Gaussian Articulated Template Models (GART)~\cite{Lei2023} is an approach for rendering and capturing from monocular videos of non-rigidly articulated subjects.} GART explicitly approximates the shape and appearance of a deformable subject by using a mixture of moving 3D Gaussians. Hu et al. introduced \textbf{GauHuman} in~\cite{Hu2023gauhuman}, a method that uses   Gaussian Splatting in canonical space and transforms 3D Gaussians to posed space using linear blend skinning (LBS). This approach incorporates effective pose and LBS refinement modules to learn fine details of 3D humans at minimal computational cost. To expedite optimization, the authors initialize and prune 3D Gaussians with a 3D human prior, employ splitting/cloning guided by KL divergence, and introduce a novel merge operation.

Abdal et al. introduced \textbf{Gaussian Shell Maps (GSMs)} in~\cite{Abdal2023shell} as a framework that connects SOTA generator network architectures with emerging 3D Gaussian rendering primitives using an articulable multi shell–based scaffold. In this setting, a CNN generates a 3D texture stack with features that are mapped to the shells. The latter represent inflated and deflated versions of a template surface of a digital human in a canonical body pose. Instead of rasterizing the shells directly, the authors sample 3D Gaussians on the shells whose attributes are encoded in the texture features.

\subsubsection{Animatable}
\label{tree:3_2}
These approaches typically train pose-dependent Gaussian maps to capture intricate dynamic appearances, including finer details in clothing, resulting in high-quality avatars. Some of these methods also support real-time rendering capabilities.\\

Jiang et al. came up with \textbf{HiFi4G in~\cite{Jiang2023}, this method efficiently renders a realistic human. HiFi4G combines 3D Gaussian representation with non-rigid tracking, employing a dual-graph mechanism for motion priors and a 4D Gaussian optimization with adaptive spatial-temporal regularizers.} They achieved a compression rate of approximately 25 times and requiring less than 2MB of storage per frame, HiFi4G excels in optimization speed, rendering quality, and storage overhead, as shown in Figure \ref{fig:hifi4g}. It presents a compact 4D Gaussian representation bridging Gaussian Splatting and non-rigid tracking. However, dependencies on segmentation, sensitivity to poor segmentation that causes artifacts, and the need for per-frame reconstruction and mesh tracking pose limitations. Future research may focus on accelerating optimization processes and reducing GPU sorting dependencies for broader deployment on web viewers and mobile devices.
\Figure[hbt!](topskip=0pt, botskip=0pt, midskip=0pt)[width=0.99\columnwidth]{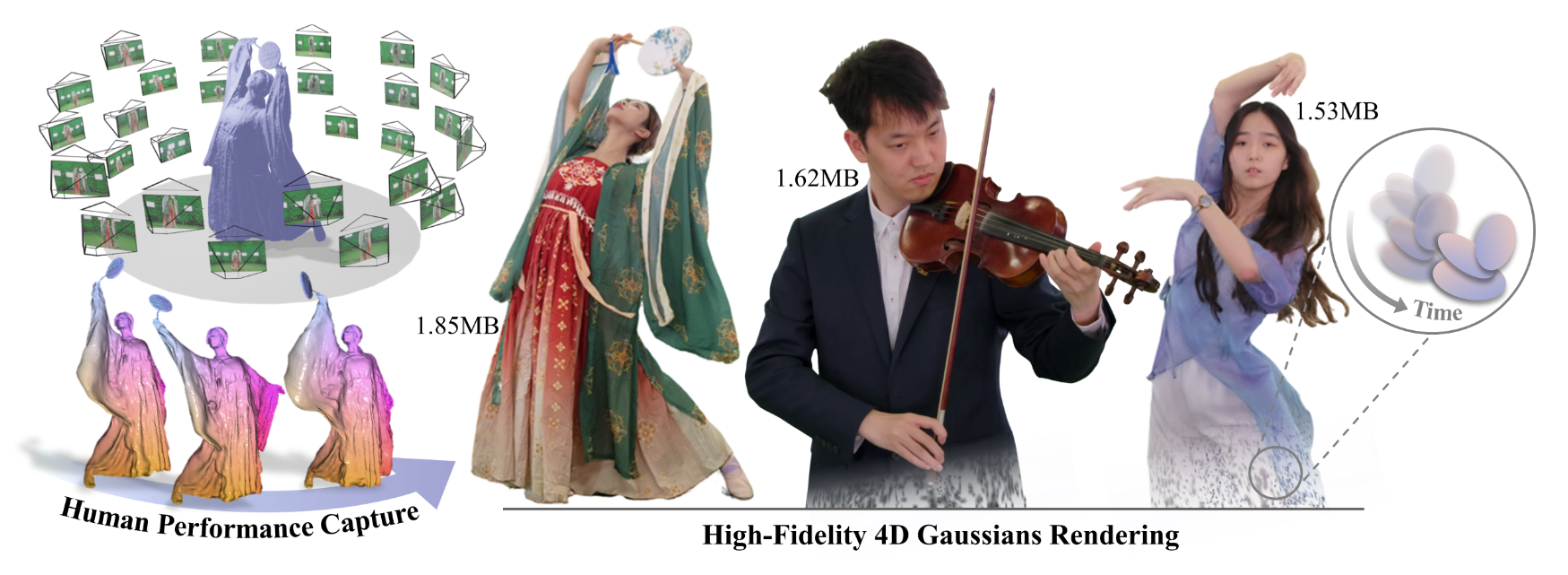}
{\textbf{4D Gaussian rendering from HiFi4G~\cite{Jiang2023}.}\label{fig:hifi4g}}

Li et al. in~\cite{Li2023animatable} leverages powerful 2D CNNs that learn a \textbf{parametric template from the input videos} which is adaptive to the wearing garments for modeling looser clothes like dresses. The authors employ a powerful StyleGAN-based CNN to learn the pose-dependent Gaussian maps for modeling detailed dynamic appearances. Very similar notable works are by Moreau et al.\cite{Moreau2023}, Kocabas et al.~\cite{Kocabas2023} and Pang et al.~\cite{Pang2023}. Zheng et al. in~\cite{Zheng2023} proposes a \textbf{fully differentiable framework composed of an iterative depth estimation module} and a Gaussian parameter regression module. The intermediate predicted depth map bridges the two components and makes them promote mutually. Also, the authors developed a real-time NVS system that achieves 2K-resolution rendering by directly regressing Gaussian parameter maps. Hu et al. in~\cite{Hu2023GaussianAvatar} came up with an  approach leveraging \textbf{differentiable motion conditions, enabling the joint optimization of motions and appearances during avatar modeling.} This addresses the persistent challenge of inaccurate motion estimation in single view settings.

\subsubsection{Head based}
\label{tree:3_3}
Previous head avatar methods have mostly relied on fixed explicit primitives (mesh, point) or implicit surfaces (SDFs). Gaussian Splatting based models will pave the way for AR/VR and the rise of filter based application, that let the user try on different makeup, shades, hairstyles etc.\\
Wang et al. leveraged canonical gaussians in~\cite{Wang2023}, to represent dynamic scenes. Using \textbf{explicit "dynamic" tri-plane as an efficient container for parameterized head geometry}, aligned well with factors in the underlying geometry and triplane, the authors obtain aligned canonical factors for the canonical Gaussians. With a tiny MLP, factors are decoded into opacity and spherical harmonic coefficients of 3D Gaussian primitives. \textbf{Quin et al.} in~\cite{Qian2023} created hyper-realistic head avatars with controllable view, pose, and expression. During avatar reconstruction, the author optimize morphable model parameters and Gaussian splat parameters simultaneously. The work demonstrated the animation capabilities of avatar in various challenging scenarios. Dhamo et al. proposes \textbf{HeadGaS}~\cite{Dhamo2023}, a hybrid model extending the explicit representation from 3D-GS with a base of learnable latent features. These features can then be linearly blended with low-dimensional parameters from parametric head models to derive expression-dependent final color and opacity values. Some example images are shown if Figure \ref{fig:HeadGaS}
\Figure[hbt!](topskip=0pt, botskip=0pt, midskip=0pt)[width=0.99\columnwidth]{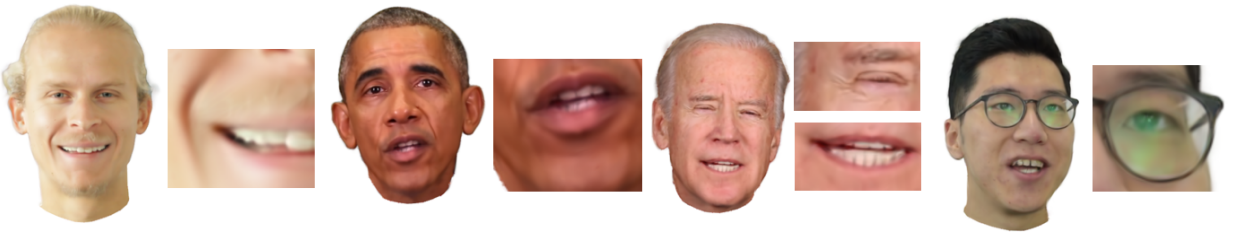}
{\textbf{HeadGaS framework generating realistic head avatars~\cite{Dhamo2023}.}\label{fig:HeadGaS}}

To model the finer facial details and expression Xiang et al. proposed \textbf{FlashAvatar} in ~\cite{Xiang2023} that uses geometric priors, and used an initialization technique for reducing number of Gaussian. Saito et al.~\cite{Saito2023} proposed \textbf{relighting of faced model}, and Chen et al. in~\cite{Chen2023mono} introduced head avatars characterized by Gaussian points \textbf{with adaptable shapes, allowing for flexible topology.} These points undergo movement according to a Gaussian deformation field aligned with the target pose and expression, enabling efficient deformation.


\subsection{Simultaneous Localization and Mapping}
\label{tree:4}
SLAM is a technique employed in autonomous vehicles to concurrently build a map and determine the vehicle's location within that map. It enables vehicles to navigate and map unknown environments. Visual SLAM (vSLAM), as the name implies, relies on images from cameras and various image sensors. This approach accommodates diverse camera types, including simple, compound eye, and RGB-D cameras, making it a cost-effective solution. Landmark detection, facilitated by cameras, can be combined with graph-based optimization, enhancing flexibility in SLAM implementation. Monocular SLAM, a subset of vSLAM using a single camera, faces challenges in depth perception, which can be addressed by incorporating additional sensors like encoders for odometry and inertial measurement units (IMUs). Key technologies related to vSLAM encompass SFM, visual odometry, and bundle adjustment. Visual SLAM algorithms fall into two main categories: sparse methods, employing feature point matching (e.g., Parallel Tracking and Mapping~\cite{PTAM}, ORB-SLAM~\cite{murORB2}), and dense methods, which utilize overall image brightness (e.g., DTAM~\cite{DTAM}, LSD-SLAM~\cite{LSD-SLAM}, DSO~\cite{DSO}, SVO~\cite{SVO}).

\textbf{GS-SLAM}~\cite{yan2023nerfds} by Yan et al. a novel approach to SLAM by leveraging a 3D Gaussian representation and a differentiable splatting rasterization pipeline, achieving real-time tracking and mapping on GPU. This method shown in Figure \ref{fig:gs_slam} outperforms SOTA alternatives with a remarkable 100× faster rendering FPS and superior full-image quality. GS-SLAM strikes a balance between efficiency and accuracy by employing a real-time differentiable splatting rendering pipeline, offering accelerated map optimization and RGB-D re-rendering compared to recent SLAM methods utilizing neural implicit representations. The proposed adaptive expansion strategy dynamically adjusts the 3D Gaussian representation, efficiently reconstructing observed scene geometry and improving mapping. This dynamic approach extends beyond synthesizing static objects and contributes to reconstructing entire scenes. The coarse-to-fine camera tracking technique enhances runtime efficiency and robust pose estimation. GS-SLAM demonstrates competitive performance on datasets like Replica~\cite{replica19arxiv} and TUM-RGBD~\cite{sturm12iros}, showcasing its efficacy in both reconstruction and localization with significantly reduced time consumption. However, GS-SLAM's dependence on high-quality depth information for 3D Gaussian initialization and updates may be a limitation in environments lacking such data. Future work aims to address this challenge by designing improved optimization methods for on-the-fly updates of 3D Gaussian positions. Additionally, efforts will be directed toward optimizing memory usage for large-scale scenes through the incorporation of neural scene representations.
\Figure[hbt!](topskip=0pt, botskip=0pt, midskip=0pt)[width=0.99\columnwidth]{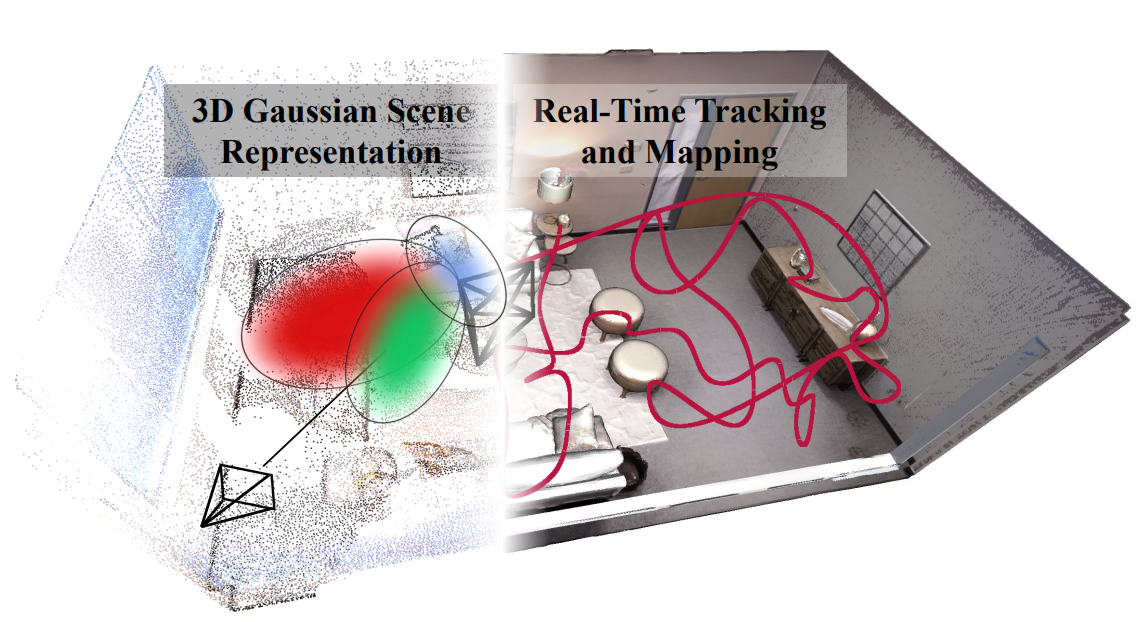}
{\textbf{GS-SLAM Framework~\cite{yan2023nerfds}.}\label{fig:gs_slam}}

\textbf{Splat, Track \& Map 3D Gaussians, SplaTAM}~\cite{keetha2023splatam} by Keetha et al. introduces an innovative approach to dense SLAM, achieving precise camera tracking and high-fidelity reconstruction in challenging real-world scenarios. This is accomplished through \textbf{online optimization of a unique volumetric representation, 3D Gaussian Splatting, utilizing differentiable rendering.} The method demonstrates sub-centimeter localization, even in texture-less environments with substantial camera motion, a challenging scenario for other state-of-the-art baselines. SplaTAM is the pioneering dense RGB-D SLAM solution using 3D Gaussian Splatting, representing the world as a set of 3D Gaussians for rendering high-fidelity color and depth images. Despite its state-of-the-art performance, the method exhibits sensitivity to motion blur, large depth noise, and aggressive rotation, prompting future exploration into temporally modeling these effects. SplaTAM's scalability to large-scale scenes through efficient representations like OpenVDB~\cite{OpenVDB} is noted, and the method currently relies on known camera intrinsics and dense depth for SLAM, suggesting future work could address reducing these dependencies. The approach achieves a remarkable rendering speed of 400 FPS for a resolution of 876×584, showcasing its efficiency in generating photo-realistic views.

The work by Matsuki et al.~\cite{matsuki2023gaussian} proposes 3D scene reconstruction using a single moving monocular or RGB-D camera. Operating at 3 FPS, the SLAM method utilizes Gaussians as the sole 3D representation, unifying accurate tracking, mapping, and high-quality rendering. Key innovations include formulating camera tracking through direct optimization against 3D Gaussians, \textbf{enabling fast and robust tracking without relying on offline SFM poses.} The explicit nature of Gaussians is leveraged for geometric verification and regularization, addressing ambiguities in incremental 3D dense reconstruction. The presented SLAM system achieves state-of-the-art results in NVS, trajectory estimation, and reconstruction of intricate details, including tiny and transparent objects, thereby significantly advancing the fidelity attainable by a live monocular SLAM system. The visual representations showcase the system's ability to capture complex material properties and details, such as transparency and thin structures, demonstrating its effectiveness in real-time 3D scene reconstruction.

Yugay et al. introduce \textbf{Gaussian-SLAM} in~\cite{Yugay2023Gaussian-SLAM}, a novel dense SLAM method incorporating Gaussian splats as a scene representation, enabling rapid, photo-realistic rendering of both real-world and synthetic scenes. This approach achieves unprecedented rendering quality, particularly evident in complex real-world datasets like TUM-RGBD frames with intricate details. \textbf{Novel strategies for seeding and optimizing Gaussian splats are proposed, facilitating their adaptation from multi-view offline scenarios to sequential monocular RGBD input setups.} The method extends Gaussian splats to encode geometry and demonstrates competitive reconstruction performance and runtime. Gaussian-SLAM outperforms existing solutions in rendering accuracy while maintaining a favorable balance of memory and compute resource usage, showcasing its efficacy for modern neural SLAM applications.

\subsection{Mesh Extraction and Physics}
\label{tree:5}
Gaussian Splatting can be used in physics based simulation and rendering. By adding more parameters in the 3D Gaussian kernel velocity, strain, and other mechanical properties can be modeled. This is why a variety of methods have been developed in the few months involving simulation of physics using Gaussian Splatting.

Xie et al. introduce a approach to \textbf{3D Gaussian kinematics based on continuum mechanics, employing Partial Differential Equations (PDEs)} in~\cite{Xie2023} to drive the evolution of Gaussian kernels and their associated spherical harmonics. This innovation allows for a unified simulation-rendering pipeline, streamlining motion generation by eliminating the need for explicit object meshing. Their method showcases versatility through comprehensive benchmarks and experiments across various materials, demonstrating real-time performance in scenes with simple dynamics. The authors presents \textbf{PhysGaussian, a framework that seamlessly generates physics-based dynamics and photo-realistic renderings simultaneously.} While acknowledging limitations such as the absence of shadow evolution in the framework and the use of one-point quadrature for volume integrals, the authors propose avenues for future work, including the adoption of high-order quadratures in the Material Point Method (MPM) and exploring the integration of neural networks for more realistic modeling. This framework can be extended to handle diverse materials like liquids and incorporate user controls leveraging advancements in Large Language Models (LLMs). Figure \ref{fig:PhysGaussian} shows the training process of PhysGaussian framework.
.;p-\Figure[hbt!](topskip=0pt, botskip=0pt, midskip=0pt)[width=0.99\columnwidth]{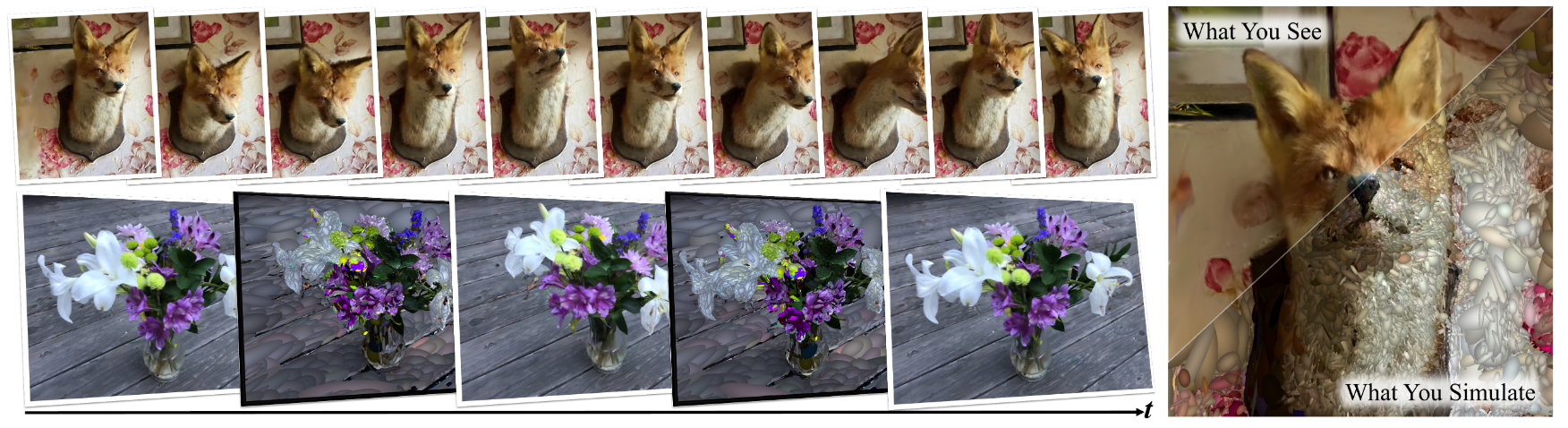}
{\textbf{PhysGaussian Framework training~\cite{Xie2023}.}\label{fig:PhysGaussian}}

Guedon et al. proposed a method~\cite{guadon2023sugar} to address the challenge of extracting a precise and rapidly obtainable mesh from the widely adopted 3D Gaussian Splatting. While Gaussian Splatting offers \textbf{realistic rendering and quicker training compared to NeRFs}, mesh extraction from the optimized and unorganized Gaussians proves difficult. The key contribution involves introducing a regularization term that encourages Gaussian alignment with the scene's surface. Leveraging this alignment, the authors present an efficient algorithm employing Poisson reconstruction for rapid and scalable mesh extraction, surpassing traditional methods like Marching Cubes. Additionally, Guedon et al. introduce an optional refinement strategy that binds Gaussians to the mesh, enabling joint optimization for seamless editing, sculpting, animation, and relighting through Gaussian Splatting rendering. This method achieves the retrieval of an editable mesh for realistic rendering within minutes, a significant improvement over state-of-the-art SDF methods that take hours, thereby enhancing rendering quality and offering versatile scene editing capabilities.

Duisterhof et al. introduces \textbf{MD-Splatting}~\cite{Duisterhof2023}. This novel approach \textbf{combines 3D tracking and NVS by leveraging video captures from multiple camera angles. MD-Splatting utilizes Gaussian Splatting, employing a deformation function based on neural-voxel encoding and a multilayer perceptron to project Gaussians into metric space.} The incorporation of physics-inspired regularization terms ensures trajectories with reduced errors. Empirical results demonstrate MD-Splatting's superior performance in simultaneous 3D metric tracking and NVS, achieving an average improvement of 16\%. The method is showcased on six synthetic scenes with large deformations, shadows, and occlusions, contributing a dataset to the research community. While highlighting its achievements, the team acknowledges the need for further exploration in real-world scenarios, considering factors like camera setup complexity and the extension to soft objects in larger environments as promising avenues for future research.

\textbf{Chen et al.} introduces a novel pipeline in~\cite{Chen2023_Neusg} that combines the strengths of \textbf{3D-GS and neural implicit models (NeuS)}~\cite{Wang2021_NeuS}. While previous methods often result in over-smoothed depth maps or sparse point clouds, the proposed approach leverages 3D Gaussian Splatting to generate dense point clouds with intricate details. To overcome challenges where the generated points may not precisely align with the surface, the paper~\cite{Chen2023_Neusg} introduces a scale regularizer to enforce thin 3D Gaussians and refines the point cloud using normal's predicted by neural implicit models. This joint optimization of 3D-GS and NeuS enhances surface reconstruction, generating complete and detailed surfaces. The empirical validation on Tanks\&Temples datasets demonstrates the effectiveness of the proposed NeuSG framework, showcasing significant improvements over previous methods in surface reconstruction quality.

\subsection{Editable}
\label{tree:7}
Gaussian Splatting has also extend it's wings to 3D editing and point manipulation of a scene. Even prompt based 3D editing of a scene is possible using the latest advancement that will be discussed. These methods not only represent the scene as 3D Gaussians, but also have a semantic and contectual understanding of the scene.

In the study~\cite{Chen2023_Swift} Chen et al. introduces \textbf{GaussianEditor, a novel 3D editing algorithm based on Gaussian Splatting, designed to overcome the limitations of traditional 3D editing methods. While conventional methods relying on meshes or point clouds struggle with realistic depiction, implicit 3D representations like NeRF face challenges related to slow processing speeds and limited control.} GaussianEditor addresses these issues by leveraging 3D-GS, enhancing precision and control through Gaussian semantic tracing and introducing Hierarchical Gaussian Splatting (HGS) for stabilized and refined results under generative guidance. The algorithm includes a specialized 3D inpainting approach for efficient object removal and integration, showcasing superior control, efficacy, and rapid performance in extensive experiments. Figure \ref{fig:gaussian_editor} shows various text promts tested by Chen et al. GaussianEditor marks a significant advancement in 3D editing, offering enhanced effectiveness, speed, and controllability. The contributions of the study include the introduction of Gaussian semantic tracing for detailed editing control, the proposal of HGS for stable convergence under generative guidance, the development of a 3D inpainting algorithm for swift object removal and addition, and extensive experiments demonstrating the method's superiority over previous 3D editing approaches. Despite its advancements, GaussianEditor relies on 2D diffusion models for effective supervision, posing limitations in addressing complex prompts, a common challenge shared with other 3D editing methods based on similar models.
\Figure[hbt!](topskip=0pt, botskip=0pt, midskip=0pt)[width=0.99\columnwidth]{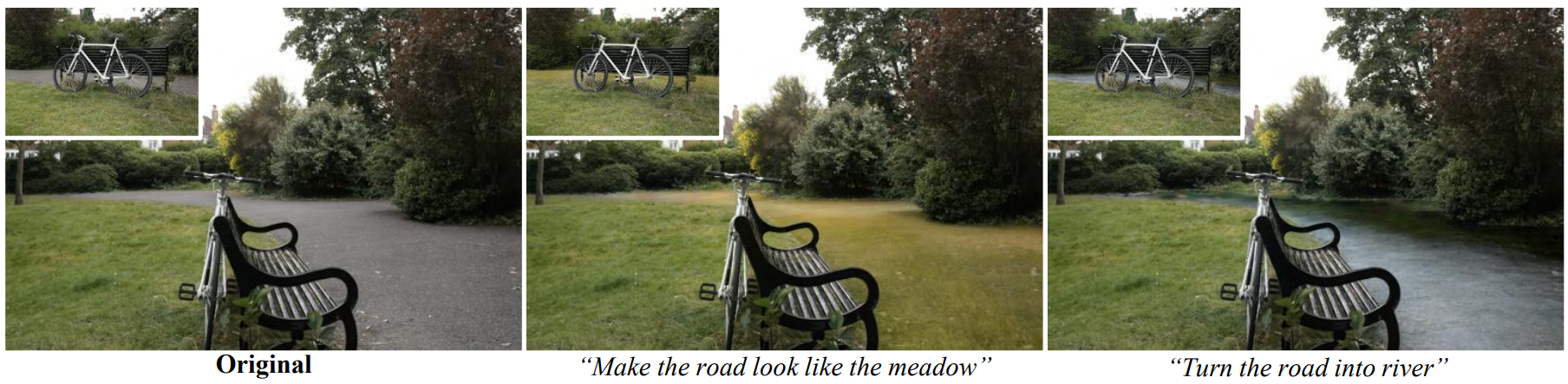}
{\textbf{GaussianEditor changing scenes with various text prompts~\cite{Chen2023_Swift}.}\label{fig:gaussian_editor}}

In the study~\cite{Fang2023} Fang et al. introduces a \textbf{systematic framework designed for delicate 3D scene editing based on 3D Gaussian Splatting, addressing limitations of current diffusion models. Unlike existing methods, this method enables precise and localized editing for 3D scenes by leveraging the explicit properties of 3D Gaussians.} The framework extracts the region of interest (RoI) from text instructions, aligns it with 3D Gaussians, and utilizes the Gaussian RoI for control during the editing process. GaussianEditor achieves more accurate and exquisite editing results compared to previous methods, such as Instruct-NeRF2NeRF, while boasting faster training speeds, completing within 20 minutes on a single V100 GPU. The contributions include being one of the first systematic methods for delicate 3D scene editing based on 3D Gaussian Splatting, proposing techniques for precise RoI localization, and achieving superior editing results with significantly reduced training time. However, some challenges persist, such as discrepancies in scene description generation from different views and potential difficulties in scenes where grounding segmentation or diffusion models fail. Future optimizations will target these issues, and the paper suggests extending Gaussian Splatting to dynamic scenes as a potential avenue for future work.

Huang et al. present \textbf{Point'n Move in~\cite{Huang2023-pm}, a technique that utilizes 3D-GS for interactive object manipulation featuring exposed region inpainting.} The method facilitates intuitive object selection, high-quality inpainting, and real-time editing through a dual-stage self-prompting mask propagation process. Despite its effectiveness in both forward-facing and $360 \degree$ scenes, limitations include a focus on geometry editing without handling lighting or texture, and potential darkening of inpaintings due to precision issues in segmentation.

In~\cite{Ye2023}, Yi et al. extend Gaussian Splatting with \textbf{Gaussian Grouping, addressing limitations in appearance and geometry modeling.} Identity Encoding facilitates object instance or stuff membership grouping, allowing efficient segmentation supervised by 2D mask predictions. Compared to implicit NeRF representations, Gaussian Grouping demonstrates reconstructive, segmentative, and editable capabilities in 3D scenes. A local Gaussian Editing scheme showcases effectiveness in diverse scene editing applications.

Cen et al. introduce \textbf{Segment Any 3D Gaussians (SAGA)} in~\cite{Cen2023} for interactive 3D segmentation, integrating 2D segmentation results into 3D Gaussians. Achieving nearly 1000× acceleration compared to previous state-of-the-art, SAGA provides real-time multi-granularity segmentation, accommodating prompts like points, scribbles, and 2D masks. Challenges include ambiguity in 3DGS-learned Gaussians and noise in SAM-extracted masks, suggesting areas for future improvements.

Zou et al. integrate 3D Gaussian Splatting with \textbf{feature field distillation}, advancing 3D scene representation for semantic tasks~\cite{Zhou2023-ff}. The framework achieves notable efficiency gains, being up to 2.7× faster than NeRF-based approaches. Experimental results demonstrate improved mIoU for semantic segmentation tasks and introduce novel capabilities like point and bounding-box prompting. While acknowledging limitations, the research signifies a significant step forward in explicit 3D feature field representation for interactive and semantically enhanced 3D scene applications.

\section{Discussion}
Traditionally, 3D scenes have been represented using meshes and points due to their explicit nature and compatibility with rapid GPU/CUDA-based rasterization. However, recent advancements like NeRF methods settles on for continuous scene representations, employing techniques such as multi layerd perceptron optimization through volumetric ray-marching for novel view synthesis. While continuous representations aid optimization, the stochastic sampling necessary for rendering introduces costly noise. Gaussian Splatting bridges this gap by leveraging a 3D Gaussian representation for optimization, achieving state-of-the-art visual quality and competitive training times. Additionally, a tile-based splatting solution ensures real-time rendering with top-tier quality. Gaussian Splatting has delivered some of best results in term of quality and efficiency while rendering 3D scenes.

Gaussian Splatting has evolved to handle dynamic and deformable objects by modifying its original representation. This involves incorporating parameters like 3D position, rotation, scaling factor, and spherical harmonics coefficients for color and opacity. Recent progress in this domain includes the introduction of a sparsity loss to encourage basis trajectory sharing, a dual-domain deformation model to capture time-dependent residuals, and Gaussian Shell Maps linking generator networks with 3D Gaussian rendering. Efforts have also been made to address challenges such as non-rigid tracking, avatar expression variation, and rendering realistic human performance efficiently. These advancements collectively aim for real-time rendering, optimized efficiency, and high-quality outcomes when dealing with dynamic and deformable objects.

In other aspect, diffusion and Gaussian Splatting synergize to create 3D objects from text prompts. Diffusion models, a type of neural network, learn to generate images from noisy inputs by reversing the process of image corruption through a sequence of increasingly clean images. In the text-to-3D pipeline, a diffusion model generates an initial 3D point cloud from a text description, which is then transformed into Gaussian spheres using Gaussian Splatting. Rendered Gaussian spheres produce the final 3D object image. Advances in this field include using structured noise to tackle multi-view geometric challenges, introducing a variational Gaussian Splatting model to address convergence issues, and optimizing denoising scores for enhanced diffusion priors, aiming for unparalleled realism and performance in text-based 3D generation.

Gaussian Splatting has been extensively applied to the creation of digital avatars for AR/VR applications. This involves capturing subjects from a minimal number of viewpoints and constructing 3D models. The technique has been used to model human body articulation, joint angles, and other parameters, enabling the generation of expressive and controllable avatars. Advancements in this area include the development of methods to capture high-frequency facial details, preserve exaggerated expressions, and efficiently deform avatars. Additionally, hybrid models have been proposed, combining explicit representations with learnable latent features to achieve expression-dependent final color and opacity values. These advancements aim to enhance the geometry and texture of generated 3D models, catering to the increasing demand for realistic and controllable avatars in AR/VR applications.

Gaussian Splatting also finds versatile applications in SLAM, providing real-time tracking and mapping capabilities on GPUs. By employing a 3D Gaussian representation and a differentiable splatting rasterization pipeline, it achieves swift and photorealistic rendering of both real-world and synthetic scenes. The technique extends to mesh extraction and physics-based simulation, allowing for the modeling of mechanical properties without explicit object meshing. Advancements in continuum mechanics and PDEs have enabled the evolution of Gaussian kernels, streamlining motion generation. Notably, optimizations involve efficient data structures like OpenVDB, regularization terms for alignment, and physics-inspired terms for reduced errors, enhancing the overall efficiency and accuracy. Other works have been done on compression, and improving rendering efficiency of Gaussian Splatting.

\subsection{Comparison}
From Table~\ref{table:4}, it is clear that at the time of writing, Gaussian Splatting is the closest option to real time rendering and dynamic scene representation. Occupancy network are not at all tailored for NVS use case. Photogrammetry is ideal for creating highly accurate and realistic models with a strong sense of context. NeRFs excel in generating novel views and realistic lighting effects, offering creative freedom and handling complex scenes. Gaussian Splatting shines in its real-time rendering capabilities and interactive exploration, making it suitable for dynamic applications. Each method has its niche and complements the others, offering a diverse range of tools for 3D reconstruction and visualization.
\begin{table}[h]
    \caption{Comparison of Photogrammetry, Occupancy
Network, NeRFs, and Gaussian Splatting.}
    \label{table:4}
    \begin{tabular}{|m{35pt}|m{85pt}|m{85pt}|}
    
        \hline
        \textbf{Method} & \textbf{Advantages} & \textbf{Disadvantages} \\
        \hline
        \textbf{Photogra- mmetry} & Accurate measurements, detailed surface textures, realistic context & Processing time, computational resources \\
        \hline
        \textbf{Occupancy Network} & Efficient representation, handles occlusions well, scalable & Limited to discrete occupancy information, struggle with detailed geometry \\
        \hline
        \textbf{NeRFs} & Novel view generation, realistic lighting effects, creative freedom & High training time, computational resources, accessibility \\
        \hline
        \textbf{Gaussian Splatting} & Real-time rendering, interactive exploration, accurate representations & Less photorealistic \\
        \hline
    \end{tabular}
\end{table}
\subsection{Challenges and Limitations}
Although Gaussian Splatting is a very robust technique, it have some caveats. Some of these are listed below:
\begin{enumerate}
    \item \textbf{Computational complexity:} Gaussian Splatting requires evaluating Gaussian functions for each pixel, which can be computationally intensive, especially when dealing with a large number of points or particles.
    
    \item \textbf{Memory usage:} Storing intermediate results for Gaussian Splatting, such as the weighted contributions of each point to neighboring pixels, can consume a significant amount of memory.
    
    \item \textbf{Edge artifacts:} Gaussian Splatting can produce undesirable artifacts near edges or high-contrast regions in the image, such as ringing or blurring.
    
    \item \textbf{Performance vs. accuracy trade-off:} Achieving high-quality results may require using a large kernel size or evaluating multiple Gaussian functions per pixel, which impacts performance.
    
    \item \textbf{Integration with other rendering techniques:} Integrating Gaussian Splatting with other techniques like shadow mapping or ambient occlusion while maintaining performance and visual coherence can be complex.
\end{enumerate}
\subsection{Future Directions}
Real-time 3D reconstruction techniques will enable several capabilities in computer graphics and related fields, such as interactively exploring 3D scenes or models in real time, manipulating viewpoints and objects with immediate feedback. It will also enable rendering of dynamic scenes with moving objects or changing environments in real time, enhancing realism and immersion. Real-time 3D reconstruction can be utilized in simulations and training environments, providing realistic visual feedback for virtual scenarios in fields such as automotive, aerospace, and medicine. It will also support real-time rendering of immersive AR and VR experiences, where users can interact with virtual objects or environments in real time. Overall, real-time Gaussian Splatting enhances the efficiency, interactivity, and realism of various applications in computer graphics, visualization, simulation, and immersive technologies.

\section{Conclusion}
In this paper, we discuss various functional and applicational aspects related to Gaussian Splatting for 3D reconstruction and novel view synthesis. It covers topics such as dynamic and deformation modeling, motion tracking, non-rigid-/deformable objects, expression/emotion variation, diffusion for text-based generation, denoising, optimization, avatars, animatable objects, head-based modeling, simultaneous localization and planning, mesh extraction and physics, optimization techniques, editing capabilities, rendering methods, compression, and more.

Specifically, the paper delves into the challenges and advancements in image-based 3D reconstruction, the role of learning-based methods in improving 3D shape estimation, and the potential applications and future directions of Gaussian Splatting techniques in handling dynamic scenes, interactive object manipulation, 3D segmentation, and scene editing.

Gaussian Splatting has transformative implications across diverse fields, including computer-generated imagery , VR/AR, robotics, film and animation, automotive design, retail, environmental studies, and aerospace applications. However, it is important to note that Gaussian Splatting may have limitations in terms of achieving photorealism compared to other methods such as NeRFs. Additionally, challenges related to overfitting, computational resources, and limitations in rendering quality should be considered. Despite these limitations, ongoing research and advancements in Gaussian Splatting continue to address these challenges and further improve the method's effectiveness and applicability.

\bibliographystyle{ieeetr}
\bibliography{library}

\begin{IEEEbiography}[{\includegraphics[width=1in,height=1.25in,clip,keepaspectratio]{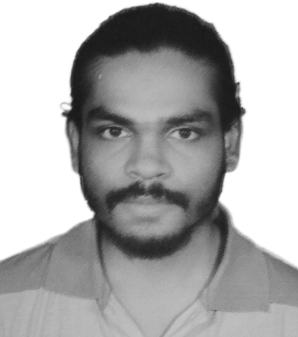}}]{Anurag Dalal} received the B.Tech in Electrical Engineering from MCKV Institute of Engineering, Howrah, West Bengal India and M.Tech in Mechatronics Engineering from Indian Institute of Engineering Science and Technology, Shibpur, West Bengal, India in 2021.

He has been working in computer vision projects after M.Tech, and in 2023 joined as a PhD candidate in University of Agder, Norway in the Department of Engineering Sciences, where he is working in 3D reconstuction.
\end{IEEEbiography}

\begin{IEEEbiography}[{\includegraphics[width=1in,height=1.25in,clip,keepaspectratio]{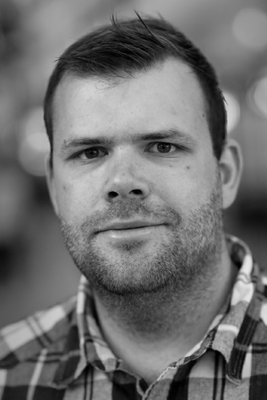}}]{Daniel Hagen} received his BSc degree in Mechatronics from the University of Agder (UiA) in 2012, after earning a trade certificate as an Automation Mechanic. He went on to complete a double MSc degree in Mechatronics through a joint program between UiA and Vorarlberg University of Applied Science (FHV) in Austria in 2014. He undertook a joint PhD program with the rig equipment company Cameron Sense AS (SLB), where he worked as a System Engineer from 2014 to 2020, and successfully defended his PhD in Mechatronics in September 2020.
While employed as a Senior R\&D Engineer, responsible for PC-based control systems of marine lifting and handling equipment, he developed a new master's course at UiA in 2021, focusing on autonomous robotics for Mechatronics students. Since 2022, he has been a full-time Associate Professor in Mechatronics, where he is involved in four master's courses related to robotics, instrumentation, programming of intelligent systems, as well as model-based design of industrial Mechatronics systems.
Currently, Daniel is the head of the national research infrastructure, Norwegian Motion Laboratory (Motion Lab), and is involved in two top-tier research centers at UiA: the Centre of Artificial Intelligence Research (CAIR) and the Top Research Centre Mechatronics (TRCM), where his research aims at developing intelligent methodologies for automated Mechatronics design and sustainable operation of autonomous offshore lifting and handling systems. He is also a work package leader in the "Legacy of Dannevig" project, which funds the work presented in this paper.
\end{IEEEbiography}

\begin{IEEEbiography}[{\includegraphics[width=1in,height=1.25in,clip,keepaspectratio]{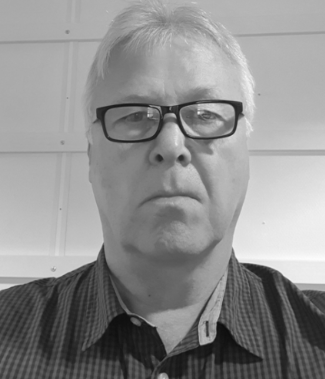}}]{Kjell G. Robbersmyr}  is a full professor in UiA – Mechatronics. Kjell has (co)authored over 130 publications in the field of mechanical engineering, vehicle crash diagnostics and CM. Kjell has long and successful leadership experience in managing large-scale research projects. In his capacity as head of research group at Agder research, Kristiansand, he was responsible for developing a test site at Lista Airport, Farsund for vehicle crash diagnostics, where around 140 tests were documented. Since 2015, Kjell heads condition monitoring group at UiA, that has published over 60 peer-reviewed journal and conference papers in the area of CM for rotating machinery making it probably the top research centre in Norway for CM. Since 2016, Kjell was a project leader on the Norwegian-Polish wind energy project (STOW) and from 2017, Kjell was a member of the centre management group of the NORCOWE program. Since 2019, Kjell is also the director of the Top Priority Research Centre in Mechatronics at UiA.  Kjell is also instrumental in realizing and conducting research under the SFI Offshore Mechatronics.
\end{IEEEbiography}

\begin{IEEEbiography}
[{\includegraphics[width=1in,height=1.25in,clip,keepaspectratio]{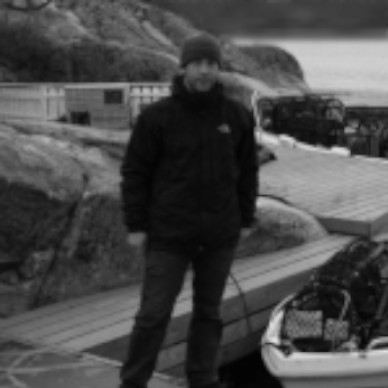}}]{Kristian Muri Knausgård} is an Assistant Professor at the University of Agder (Norway) and an IEEE Senior Member. He earned his M.Sc. in Engineering Cybernetics from NTNU in 2007. Previously serving as Assistant Director of Applications at the Top Research Centre Mechatronics (TRCM) at UiA, he has since shifted his focus to broader academic contributions in his field. His research interests include computer vision in unstructured environments, mathematical modeling, control systems, state estimation, deep learning, reinforcement learning, cybersecurity, and embedded real-time systems. He has ten years of experience from IT and industry, and has supervised more than 100 B.Sc. and M.Sc. students over the last eight years.
\end{IEEEbiography}
\EOD
\end{document}